\documentclass[journal]{IEEEtran}

\usepackage{cite}
\usepackage{amsmath,amssymb,amsfonts}

\DeclareMathOperator*{\argmin}{arg\,min}
\usepackage[noend]{algorithmic}
\usepackage{algorithm}
\usepackage{graphicx}
\usepackage{textcomp}
\usepackage{subcaption}
\usepackage{textcomp}
\usepackage{xcolor}
\usepackage{mathtools}
\usepackage{makecell}
\usepackage{dsfont}
\usepackage{hyperref}

\usepackage{array}
\newcolumntype{P}[1]{>{\centering\arraybackslash}p{#1}}

\def\etal{\emph{~et~al. }}
\makeatother

\ifCLASSINFOpdf
\else
\fi
\begin{document}
\title{Social Coordination and Altruism \protect\\ in Autonomous Driving}
% \title{Altruistic Autonomous Driving in \protect\\ Mixed Autonomy Traffic}
% \title{Inter-agent Coordination for Altruistic Autonomous Vehicles in Mixed Autonomy Traffic}

\author{Behrad Toghi$^{*}$, Rodolfo Valiente$^{*}$, Dorsa Sadigh, Ramtin Pedarsani, Yaser P. Fallah
\thanks{$^{*}$Authors B. Toghi and R. Valiente contributed equally.}
\thanks{Behrad Toghi, Rodolfo Valiente, and Yaser P. Fallah are with the Department of Electrical and Computer Engineering, University of Central Florida. {\tt\small toghi@knights.ucf.edu}}
\thanks{Dorsa Sadigh is with the Department of Electrical Engineering and the Department of Computer Science, Stanford University.}
\thanks{Ramtin Pedarsani is with the Department of Electrical and Computer Engineering, UC Santa Barbara.}
\thanks{This material is based upon work partially supported by the National Science Foundation under Grant No. CNS-1932037.}}

% \markboth{\normalfont{IEEE Transactions on Intelligent Vehicles}}%
% {Shell \MakeLowercase{\textit{et al.}}: Bare Demo of IEEEtran.cls for IEEE Journals}
\markboth{\normalfont{Under review in an IEEE Journal}}%
{Shell \MakeLowercase{\textit{et al.}}: Bare Demo of IEEEtran.cls for IEEE Journals}

\maketitle

\begin{abstract}
Despite the advances in the autonomous driving domain, autonomous vehicles (AVs) are still inefficient and limited in terms of cooperating with each other or coordinating with vehicles operated by humans. A group of autonomous and human-driven vehicles (HVs) which work together to optimize an altruistic social utility can co-exist seamlessly and assure safety and efficiency on the road. Achieving this mission without explicit coordination among agents is challenging, mainly due to the difficulty of predicting the behavior of humans with heterogeneous preferences in mixed-autonomy environments. Formally, we model an AV's maneuver planning in mixed-autonomy traffic as a partially-observable stochastic game and attempt to derive optimal policies that lead to socially-desirable outcomes using a multi-agent reinforcement learning framework (MARL), and propose a semi-sequential multi-agent training and policy dissemination algorithm for our MARL problem. We introduce a quantitative representation of the AVs' social preferences and design a distributed reward structure that induces altruism into their decision-making process. Altruistic AVs are able to form alliances, guide the traffic, and affect the behavior of the HVs to handle competitive driving scenarios. We compare egoistic AVs to our altruistic autonomous agents in a highway merging setting and demonstrate the emerging behaviors that lead to improvement in the number of successful merges and the overall traffic flow and safety.
\end{abstract}
\IEEEpeerreviewmaketitle

%
%%%%%%%%%%%%%%%%%%%%%%%%%%%%%%%%%%%%%%%%%%%%%%
%%%%%%%%%%%%%%%%%%%%%%%%%%%%%%%%%%%%%%%%%%%%%%
%%%%%%%%%%%%%%%%%%%%%%%%%%%%%%%%%%%%%%%%%%%%%%
%
\section{Introduction}
\begin{figure*}[t!]
\centering
\includegraphics[width=.99\textwidth]{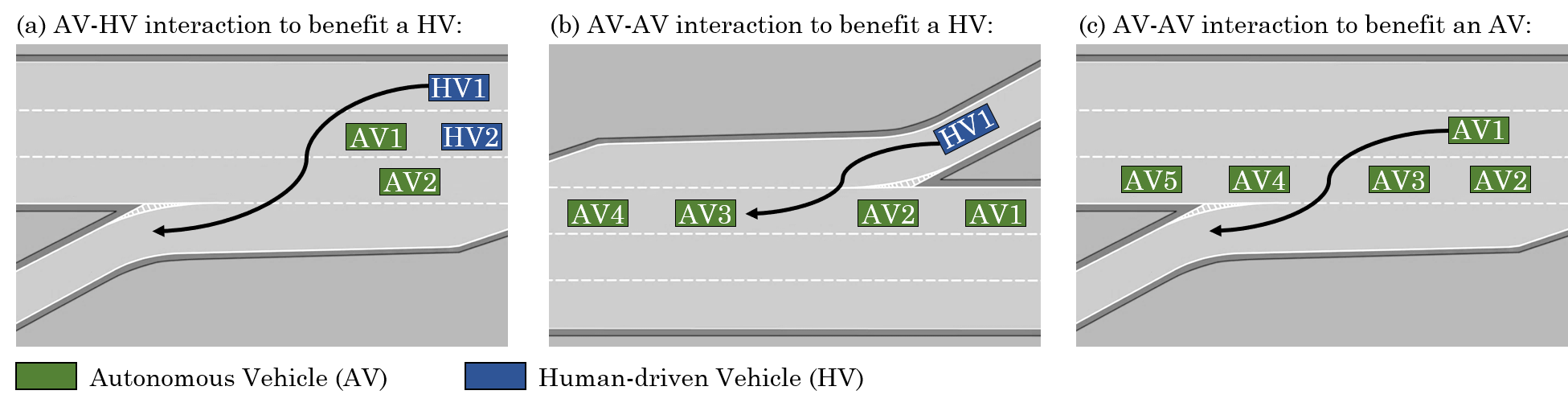}
\caption{\small{
\textbf{(a) AV-HV interaction to benefit another HV:} Altruistic agents have the opportunity to form alliances and guide the behavior of HVs in order to improve the traffic flow and avoid hazardous situations. AV1 \& AV2 can build a formation to slow down HV2 and open up a pathway for HV1, enabling it to trust the AVs, change lanes, and navigate towards the exit ramp.
\textbf{(b) AV-AV interaction to benefit another HV:} HV1 is intended to merge into the highway. Egoistic AVs ignore the merging vehicle and do not open up space for it which can potentially lead to hazardous scenarios, whereas if they show sympathy for the merging HV, they can compromise on their own interest in order to create a safe path for HV1 to merge into the highway. 
\textbf{(c) AV-AV interaction to benefit another AV:} AV1 attempts to exit the highway. If AV2-AV5 act egoistically, AV1 might miss the exit and not be able to follow its planned mission. However, if AV2-AV5 take into account the interest of AV1 and act altruistically, they can open up space in the platoon, by AV2 \& AV3 decelerating and AV4 \& AV5 accelerating, to enable a safe exit for AV1.
}}
\label{fig:mainpaperfigure}
\end{figure*}
\IEEEPARstart{C}{onnected} and automated vehicles (CAVs) pursue a mission to enhance driving safety and reliability by bringing automation and intelligence into vehicles, which lessens the inherent human limitations such as range of vision, reaction time, and distraction. Adding the communication component to intelligent vehicles further improves their ability to perceive their surroundings and creates an opportunity for mass coordination and cooperative decision-making. This inter-agent coordination is particularly important as the full potential of CAVs does not lie in operating a single vehicle on an empty road but rather from their seamless co-existence with other autonomous and human-driven vehicles (HVs). Hence, we narrow the focus of this work to studying the decision-making problem in the presence of multiple autonomous agents and human drivers, i.e. a mixed-autonomy multi-agent environment. 

% Why multi-agent coordination is hard in mixed-autonomy? what existing works do? 
Leveraging vehicle-to-vehicle (V2V) communication, decision-making in a purely-autonomous environment can be simplified into a centralized control problem with essentially one agent. However, the presence of HVs makes the inter-agent coordination more challenging as they cannot explicitly communicate to coordinate with AVs in real-time. In order to make safe and socially-desirable decisions in the presence of humans, current solutions on social navigation for AVs mainly rely on learned or hand-coded models that predict the behavior of human drivers~\cite{alahi2016social, sadigh2016planning}. We identify two key shortcomings in the existing schemes. First, the fidelity of the human models that are derived in the absence of autonomous agents is questionable in mixed-autonomy settings as human drivers tend to act differently when around AVs~\cite{sadigh2020influencing}. Second, single-agent solutions do not fully exploit the potential of CAVs in constituting a mass intelligence, forming alliances, and performing coordinated multi-agent maneuvers. 

%% what is our key insight?
We study the mixed-autonomy decision-making problem from a multi-agent point of view, as opposed to the previous individual perspectives. Our key insight is that incentivizing AVs on adopting an \emph{altruistic behavior} and accounting for the interest of other vehicles, allows them to see the big picture and find solutions that are optimal for the group in a longer term. In addition to the potential safety and efficiency benefits of altruistic decision-making, altruism leads to circumstances where no vehicle has superiority over the others, creating more societally beneficial outcomes~\cite{schwarting2019social}. To elaborate, Figure~\ref{fig:mainpaperfigure}(a) shows that a group of AVs can guide the behavior of human drivers to improve safety and efficiency, Figures~\ref{fig:mainpaperfigure}(b) and~\ref{fig:mainpaperfigure}(c) illustrate examples of how AVs can work together to achieve a social goal that benefits another HV or AV. 

% our method and contributions
We focus our work on inherently competitive driving scenarios, such as the examples illustrated in Figure~\ref{fig:mainpaperfigure}, where safe and efficient traffic flow necessarily requires coordination among autonomous agents and egoistic behavior most likely compromises on traffic safety and efficiency. 
We build on our prior work in~\cite{ toghi2021altruistic, toghi2021cooperative} and proposed a novel semi-sequential multi-agent training and policy dissemination algorithm to alleviate the non-stationary problem. Additionally, we use a method for scoring the entries in the experience replay buffer that improves sample efficiency and speeds up the learning process. Furthermore, we emphasize the importance of finding the optimal social value orientation and in contrast to the other works, formulate it as a convex optimization problem.
We formalize the mixed-autonomy driving problem as a partially observable stochastic game (POSG) and derive optimal policies using deep multi-agent reinforcement learning (MARL). With our solution, altruistic autonomous agents not only learn to drive safely but also master the inter-agent coordination and social navigation. Our main contributions are as follows:
\begin{itemize}
    \item We propose a MARL framework to train altruistic agents using a decentralized social reward signal. These agents are able to drive safely on the highway and coordinate with each other in the presence of human drivers.
    \item We proposed a novel semi-sequential multi-agent training and policy dissemination algorithm for our MARL problem and utilized a network architecture that allows our agents to implicitly learn from experience, without the need for an explicit behavioral model of human drivers. 
    \item In contrast with the existing solutions, we formulate the problem of finding the optimal social value orientation angle as a convex optimization objective. We show that an optimal value for the level of altruism exists and when chosen properly between being absolutely selfless or selfish, despite some agents' compromise on their local utility, the overall traffic safety and flow improve for the group of vehicles.
\end{itemize}

%
%%%%%%%%%%%%%%%%%%%%%%%%%%%%%%%%%%%%%%%%%%%%%%
%%%%%%%%%%%%%%%%%%%%%%%%%%%%%%%%%%%%%%%%%%%%%%
%%%%%%%%%%%%%%%%%%%%%%%%%%%%%%%%%%%%%%%%%%%%%%
%
\section{Related Work}
\label{sec:relatedworks}
This section presents a short literature review on the main topics that are closely related to our problem, namely core MARL solutions, cooperative algorithms, human behavior modeling, and navigation in the presence of humans.
\smallskip
\noindent \textbf{Multi-agent Reinforcement Learning. }
Early solutions for multi-agent value-learning algorithms assume independently trained agents and are proved to perform poorly~\cite{matignon2012independent}. To alleviate this problem, a learning rule is presented by Foerster\etal that relies on an additional term to take into account the effect of other agents' evolution during the training. They have also attempted to leverage a multi-agent derivation of importance sampling and removing outdated samples from the experience replay buffer~\cite{foerster2017stabilising} to make it effective for multi-agent settings. Xie\etal employ latent representations of partner strategies to address this problem and enable a more scalable partner modeling~\cite{xie2020learning}. Shih\etal further consider the effects of repeated interactions on partner modeling and develop a modular approach that separates rule-dependent representations from partner-dependent conventions~\cite{shih2021critical}.

Foerster\etal proposed the counterfactual multi-agent (COMA) algorithm that is expected to address the credit assignment problem in multi-agent environments~\cite{foerster2018counterfactual}. COMA algorithm utilizes the set of joint actions of all agents as well as the full state of the world during the training. In contrast, we assume partial observability and a decentralized reward function during both training and execution. More application-oriented related works, include the centralized multi-agent solutions proposed by Gupta\etal~\cite{gupta2017cooperative}. More recently, Wang\etal proposed a gifting approach that enables the emergence of prosocial behaviors in general-sum coordination games~\cite{wang2021emergent}. Importantly, in contrast with our approach, the existing literature on multi-agent systems relies on assumptions on the social preference of agents~\cite{omidshafiei2017deep, lauer2000algorithm}.
\smallskip
\noindent \textbf{Human Behavior Modeling. }
Driving styles of human drivers can be learned either from demonstration through inverse RL, as proposed by Kuderer\etal, or employing statistical models such as Gaussian and Dirichlet processes
% ~\cite{kuderer2015learning, 8690570, 8690965, mahjoub2019v2x}.
~\cite{kuderer2015learning, 8690570}.
Kuefler\etal adopt a novel approach and apply generative adversarial networks to imitate the behavior of a human driver~\cite{kuefler2017imitating}. Schmerling\etal study the scenarios with inherent multimodal uncertainty, such as our driving, and leverage conditional variational autoencoders (CVAEs) to condition the policy on the present interaction history~\cite{schmerling2018multimodal}. Recent data-driven approaches have shown achievements in classifying human driving maneuvers~\cite{toghi2020maneuver}, and predicting human trajectories to enable fully-autonomous navigation of a robot in human-dense environments~\cite{chen2017socially}. In contrast with works in the broad literature on human behavior modeling that take a game-theoretic or optimization-based approach, we rely on implicitly learning from interaction data within our MARL platform.
\smallskip
\noindent \textbf{Social Navigation. }
Alahi\etal introduced the Social LSTM framework which leverages recurrent neural networks to extract the temporal information from the trajectory of pedestrians in large crowds~\cite{alahi2016social}. Tsoi\etal present their high-fidelity simulation platform, SEAN, to accelerate the research on social robot navigation~\cite{tsoi2020sean}. Vazquez\etal study the social interactions in a human-robot role-playing game and expand their observations to studying spatial behavior of a group of robots. More recent works in social navigation have revealed the potential for collaborative planning and interaction with humans. Examples include but are not limited to works by Trautman\etal and Nikolaidis\etal where a mutual reward function is optimized in order to enable joint trajectory planning for humans and robots~\cite{trautman2010unfreezing, nikolaidis2015efficient}.
\smallskip
\noindent \textbf{Mixed-autonomy Traffic Networks. }
Lazar\etal take a more abstract and traffic-level perspective to study the emergent behaviors in mixed-autonomy environments using model-free RL solutions~\cite{lazar2019learning}. Wu\etal explore the idea of stabilizing the traffic flow that is guided by autonomous vehicles as well as the emergent behaviors in a mixed AV-HV setting~\cite{wu2018stabilizing, wu2017emergent}. Vinitsky\etal present a benchmark for traffic control based on RL in mixed-autonomy traffic~\cite{vinitsky2018benchmarks}. Biyik\etal formalize the effects of altruistic driving in mixed-autonomy at a road-level and present a formal model of road congestion that can be used for optimal routing in road networks~\cite{biyik2018altruistic}.
%

%
%%%%%%%%%%%%%%%%%%%%%%%%%%%%%%%%%%%%%%%%%%%%%%
%%%%%%%%%%%%%%%%%%%%%%%%%%%%%%%%%%%%%%%%%%%%%%
%%%%%%%%%%%%%%%%%%%%%%%%%%%%%%%%%%%%%%%%%%%%%%
%
\section{Preliminaries}
\label{sec:preliminaries}
In this section, we provide the preliminary concepts that are essential in the following section and introduce our formal notation.

\smallskip
\noindent \textbf{Partially-observable Stochastic Games.}
Decision-making process in a finite set of autonomous agents $\mathcal{I}$ with partial observability in stochastic environments can be formalized as a partially-observable stochastic game (POSG) defined by the tuple $\mathcal{M}_\text{G} \coloneqq (\mathcal{I}, \mathcal{S}, [ \mathcal{A}_i ], [ \mathcal{O}_i ], T, [ R_i ])$ for $i=1, ..., N$. At a given time, each agent receives a local observation $\textbf{o}_i: \mathcal{S} \rightarrow \mathcal{O}_i$ that is correlated with the underlying state of the environment $s \in \mathcal{S}$ and takes an action from the action space $a \in \mathcal{A}$. Consequently, the environment evolves to a new state $s'_i$ with probability $T=\Pr(s'|s, a): \mathcal{S} \times \mathcal{A}_1 \times ... \times \mathcal{A}_N \rightarrow \mathcal{S} $ and the agent receives a decentralized reward $R_i: \mathcal{S} \times \mathcal{A}_i \rightarrow \mathbb{R}$. The probability distribution over actions at a given state is known as the stochastic policy $\pi_i: \mathcal{O}_i \times \mathcal{A}_i \rightarrow [0, 1]$. The goal is to derive a distribution that maximizes the discounted sum of future rewards over an infinite time horizon, i.e., an optimal policy $\pi^*: \mathcal{S} \to \mathcal{A}$,
\begin{equation}
\label{equ:optimalpolicy}
\pi^* \coloneqq \underset{\pi}{\arg\max} \; \mathbb{E} \Big( \sum_{i=0}^{\infty} \gamma^i R\big(s_i,\pi(s_i)\big) \Big)
\end{equation}
in which, $\gamma \in [0,1)$ is the discount factor. The optimal policy maximizes the state-action value function, i.e., $\pi^*(s) = \arg\max_a Q^* (s,a)$, where
\begin{equation}
\label{equ:qfunction}
Q^\pi(s,a) \coloneqq \mathbb{E} \Big(\sum_{i=1}^\infty \gamma^i R\big(s_i, \pi (s_i)\big) |s_0=s, a_0=a \Big)
\end{equation}
and the optimal state-action value function can then be derived using the Bellman optimality equation,
\begin{equation}
\label{equ:bellmanequ}
Q^*(s,a) = \mathbb{E}_{s'\sim P(.|s,a)} \Big( R(s,a) + \max_{a'} \gamma Q^*(s',a') \Big)
\end{equation}
\smallskip
\noindent \textbf{Solving POSGs with Unknown Dynamics.}
Dynamics of the environment and reward function are usually stochastic and not fully-known in real-world problems. Reinforcement learning (RL) provides a possibility to solve POSGs with unknown reward and state transition functions through continuous interaction with the environment. RL algorithms such as off-policy temporal difference learning enable agents to update the value function from such interactions with the environment,
\begin{multline}
\label{equ:TDlearning}
Q_{i+1}(s,a) - Q_i(s,a) =\\
\alpha_i \Big( R\big(s, \pi (s)\big) + \gamma \max_{a'} Q_i(s',a') - Q_i(s,a) \Big),
\end{multline}
where $\alpha_i$ is the learning rate at the $i$th iteration. 

\smallskip
\noindent \textbf{Deep Q-networks.}
Parameterizing the state-action value function using a function approximator, i.e., $\Tilde{Q}(.;\textbf{w}) \cong Q(.)$, results in more generalizable policies that can scale to larger state-spaces. Parameters $\textbf{w}$ can be learned through mini-batch gradient descent steps,
\begin{equation}
\label{equ:Qlearning}
\textbf{w}_{i+1} = \textbf{w}_i + \alpha_i \hat{\nabla}_\textbf{w} \mathcal{L}(\textbf{w}_i)\\
\end{equation}
where, the $\hat{\nabla}_\textbf{w}$ operator estimates the gradient at $\textbf{w}_i$. Deep neural networks are widely used as function approximators and are also applicable to the Q-learning algorithm~\cite{mnih2013playing}. A deep Q-network (DQN) builds up on two major ideas, namely using two separate networks during training and employing an experience replay buffer to decorrelate the training samples. The former is done to stabilize the training process by updating the greedy network at each training iteration to compute the optimal Q-value and using another less-frequently updated target network. The loss function in Eq.~\eqref{equ:Qlearning} can be written as
\begin{equation}
\label{equ:GD_loss_target}
\mathcal{L}(\textbf{w}_i) = \mathbb{E} \Big(R+\gamma \underset{a'}{\max} \widetilde{Q}^*(s',a';\hat{\textbf{w}}) - \widetilde{Q}^*(s,a;\textbf{w})\Big)^2
\end{equation}
where $\hat{\textbf{w}}$ is the target network which periodically gets updated during the training. Additionally, DQN algorithm draws batches of training data $(s, a, R, s')$ from an experience replay buffer in order to decorrelate the training samples in Eq.~\eqref{equ:Qlearning} that are generated from simulation or real-world experience and thus naturally have temporal dependencies. This process is challenging in MARL since, $\Pr(s'|s, a, \pi_1,...,\pi_n) \neq \Pr(s'|s, a, \pi'_1,...,\pi'_n)$ if any $\pi_i \neq \pi'_i$. In other words, the environment becomes non-stationary when multiple agents are evolving concurrently. We will further discuss this issue and provide a solution to stabilize the multi-agent learning process in Section~\ref{sec:DMARL}.

\smallskip
\noindent \textbf{V2V Networks.}
We are interested in a multi-agent setting where agents have no information about others' actions and cannot explicitly coordinate. Instead, the decentralized coordination among agents is expected to arise from the social reward signal. We extend the earlier introduced concepts to a coordinated POSG defined as $(\mathcal{I}, \mathcal{S}, [ \mathcal{A}_i ], [ \widetilde{\mathcal{O}}_i ], T, [ R_i ], \mathcal{G})$,  where $\mathcal{G}=(\mathcal{I}, \mathcal{E})$ is a stochastic, time-varying, undirected graph that encompasses the V2V communication among agents in the environment $\mathcal{E}$. The communicated information can be as simple as kinematics information, e.g., speed, location, heading, or more bandwidth-intensive forms of sensory data, e.g., camera and LiDAR. Leveraging this shared situational awareness, agents can extend their range of perception and overcome obstacles and line-of-sight visibility limitations~\cite{emad2020feature, valiente2019controlling}. An agent's local observation $\Tilde{\textbf{o}}_i \in \widetilde{\mathcal{O}}_i$ is created using the shared situational awareness and clearly depends on $\mathcal{G}$ which incorporates the flow of information among agents. We utilize the network analysis from
% ~\cite{toghi2018multiple, toghi2019analysis, toghi2019spatio, saifuddin2020performance}
~\cite{toghi2018multiple}
to model the V2V communication in a high-density highway.

%
%%%%%%%%%%%%%%%%%%%%%%%%%%%%%%%%%%%%%%%%%%%%%%
%%%%%%%%%%%%%%%%%%%%%%%%%%%%%%%%%%%%%%%%%%%%%%
%%%%%%%%%%%%%%%%%%%%%%%%%%%%%%%%%%%%%%%%%%%%%%
%
\section{Problem Statement}
\label{sec:problem_statement}
We investigate the maneuver-level decision-making problem for AVs to explore behaviors that can lead to socially-desirable outcomes. We are interested in the question of how autonomous agents can be trained from scratch to perform an individual task such as driving safely on a road, while considering the social aspects of their mission, i.e., optimizing for a social utility that also accounts for the interest of other vehicles around them. Figure~\ref{fig:mainpaperfigure} helps us to build more intuition on the topic by depicting instances of driving scenarios in which altruism leads to socially-valuable outcomes and clearly overcomes the limitations of egoistic and single-agent planning. Each example in Figure~\ref{fig:mainpaperfigure} provides an example on altruistic inter-agent coordination settings that can benefit both HVs and AVs. It is clear that in some instances, altruistic AVs have to compromise on their individual utility, e.g., by slowing down, in order to increase the group's overall utility. The balance between an AV's selflessness and selfishness is the key to reaching efficient and safe traffic flow. In~\cite{ toghi2021altruistic, toghi2021cooperative}  we show that tuning the level of altruism in AVs leads to different emerging behaviors and affects the traffic flow and driving safety. In this work, we further explore that finding and formulate the problem as a convex optimization objective, to obtain an optimal social value orientation angle. Thus, we continue this section by providing a quantitative representation of an agent's level of altruism and formally defining our case study scenario, before presenting our proposed solution in the next section.

\subsection{Quantifying Social Value Orientation}
In order to formally study the social dilemmas between humans and autonomous agents in heterogeneous environments, it is crucial to quantify the social preference of an individual, e.g., whether if they will defect or cooperate in a given situation such as opening a gap in our highway merging example. The degree of an agent's egoism or altruism with regards to its counterparts is defined as \textit{Social Value Orientation (SVO)}, a widely used notion in the social psychology literature which has been recently adopted in robotics research. Specifically, we borrow the angular annotation for SVO as defined by Liebrand~\etal~\cite{liebrand1988ring}. The SVO angular preference $\phi$, quantifies how an agent weights its own reward against the reward of others. An agent's total utility $R_i$ can then be written as, 
\begin{equation}
\label{equ:svodefinition}
R_i = r_i \cos \phi_i + r^-_i \sin \phi_i
\end{equation}
where $r_i$ is the agent's individual utility, $r^-_i$ is the total utility of other agents from the perspective of the $i$th agent which in general is a function $f(.)$ of their individual utilities,
\begin{equation}
\label{equ:othersutility}
r^-_i = f(r_j), \quad \text{where } j \neq i 
\end{equation}
\begin{figure}[t]
\centering
\includegraphics[width=.45\textwidth]{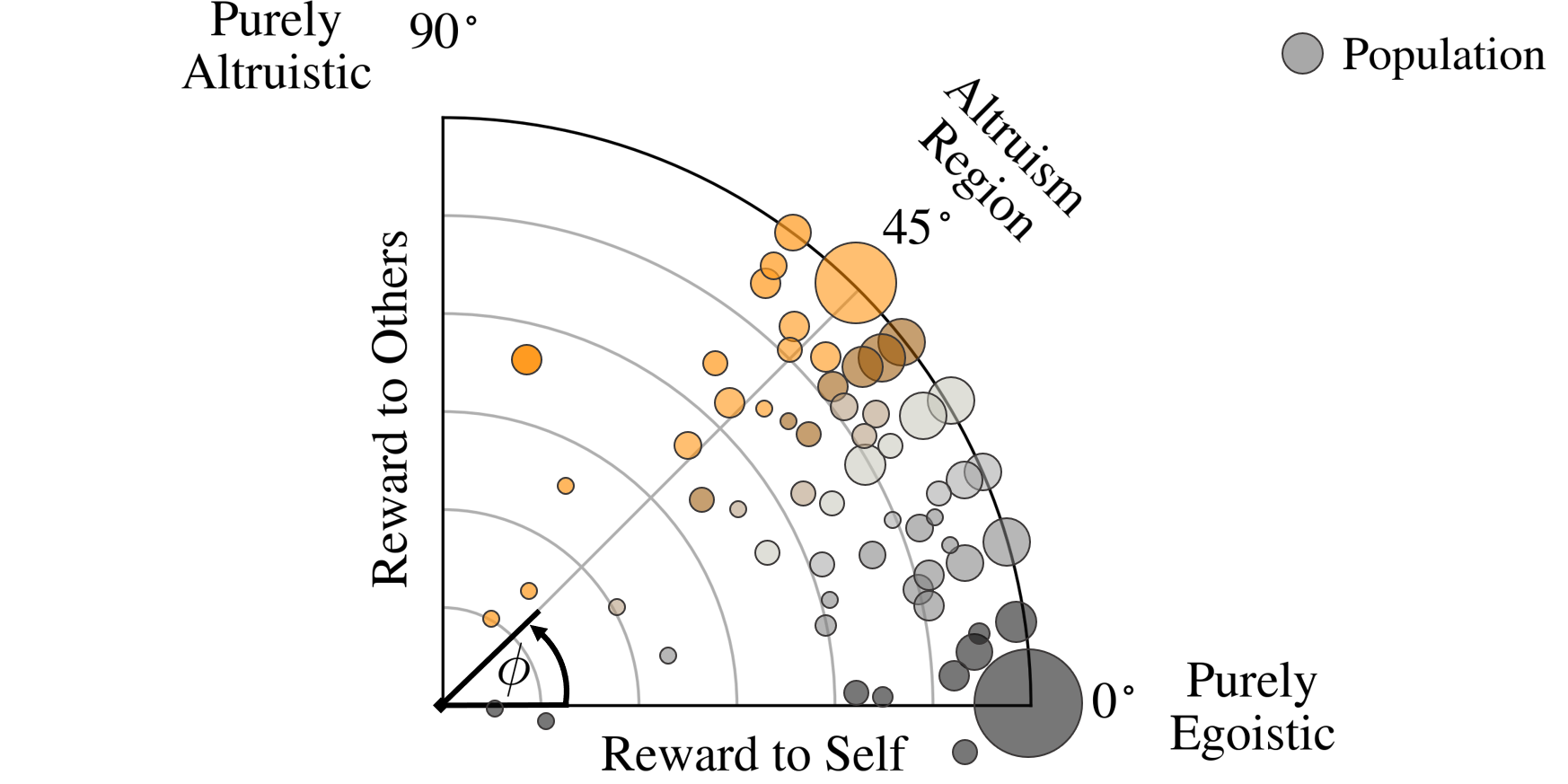}
\caption{\small{SVO angular phase $\phi$ quantifies an agent's level of altruism. Figure is based on the empirical data collected from humans by Garapin\etal\cite{garapin2015does}. Diameter of the circles show the size of the human population that hold the corresponding SVO.}}
\label{fig:svoring}
\end{figure}

Autonomous agents require an understanding of human drivers' social preferences and their willingness to coordinate. However, it is well-established in the behavioral decision theory that humans are heterogeneous in SVO and thus their preference is rather ambiguous and unclear~\cite{murphy2015social}. Current works on social navigation for AVs often make restrictive assumptions on human drivers' social preference and compliance~\cite{sadigh2016planning}, whereas Figure~\ref{fig:svoring} indicates an spectrum of altruism among humans with heterogeneous social value orientations. Thus, due to the large spectrum of altruistic behavior observed by humans, our insight is to rely on autonomous cars instead to guide the overall system toward more socially desirable objectives. Specifically, we plan to find policies for AVs that improve the utility of the group as a whole through emerging alliances and more importantly, affecting the behavior of human drivers. In our particular driving example, the desired social outcome is achieving seamless and safe highway merging while maximizing the distance traveled by all vehicles and avoiding collisions.

\subsection{Formalism}

\begin{figure}[t]
  \centering
  \includegraphics[width=.47\textwidth]{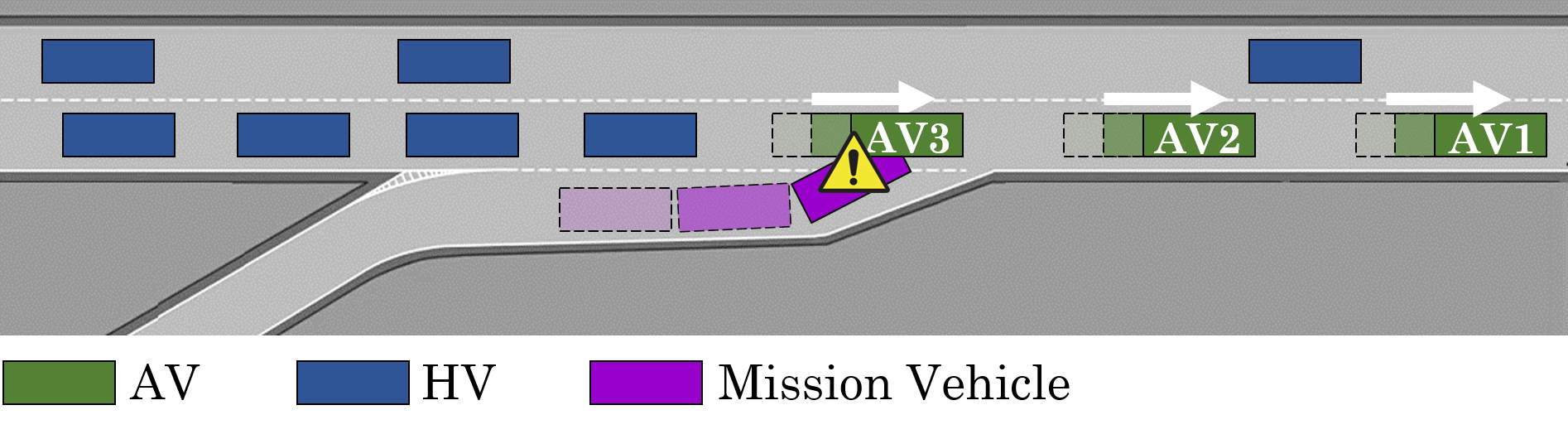}
  \caption{\small{Case study: a mission vehicle that can be human-driven or autonomous attempts to merge into the mixed group of AVs and HVs.}}
  \label{fig:formalscenario}
\end{figure}

We choose a highway merging scenario with a mixed group of AVs and HVs as our base experiment scenario, as illustrated in Figure~\ref{fig:formalscenario}. A merging vehicle, which can be either HV or AV, approaches the highway on the merging ramp and faces a mixed platoon of vehicles that are cruising on the highway. This configuration contains a group of AVs that hold the same SVO, as well as a group of HVs which are heterogeneous in their SVO and hence it is unclear if they are allies or foes. In this settings, it is obvious that the individual interest of the merging vehicle, i.e., seamless merging into the highway, does not align with that of the cruising vehicles, i.e., cruising with optimal speed and energy consumption. We design our case study scenario in a way that safe and seamless merging necessarily requires all AVs to work together and none of them alone can enable the merging of the mission vehicle without the cooperation of the others. Formally, the road section shown in Figure ~\ref{fig:formalscenario} is shared by a set of AVs $\mathcal{I}$ that are connected together via V2V communication and governed by a decentralized stochastic policy, a set of HVs $\mathcal{V}$ operated by humans with heterogeneous and unknown SVOs, and a human-driven or autonomous \textit{mission vehicle} $M \in \mathcal{I} \cup \mathcal{V}$ that attempts to merge into the highway.

A human driver's perception is often limited by their range of vision, occlusion, and obstacles. In contrast, CAVs share their observations to overcome these limitations. Each CAVs has a unique local observation $\Tilde{\textbf{o}}_i([\textbf{o}_i];\mathcal{G})$ that is constructed using the its own local observation, as well as the local observations it receives from the neighboring CAVs. As mentioned before, graph $\mathcal{G}$ grasps this inter-agent communication. Therefore, an observer AV can detect a subset of other AVs, $\widetilde{\mathcal{I}} \subset \mathcal{I}$, and a subset of HVs $\widetilde{\mathcal{V}} \subset \mathcal{V}$. As we elaborated before, our aim is to find a decentralized control scheme that can induce altruism in the behavior of AVs. Hence, each AV must use its local observation $\textbf{o}_i$ to make independent decisions that optimize its utility. The value of the agent's altruism, i.e., the SVO angular phase $\phi$, determines the social implications of an agent's local actions. To summarize, we state our problem as \textit{deriving a utility function that enables the AVs to handle competitive driving scenarios, such as those illustrated in Figure~\ref{fig:mainpaperfigure}, and lead them into socially-desirable outcomes that improve traffic safety and efficiency for the group of vehicles}. 

%
%%%%%%%%%%%%%%%%%%%%%%%%%%%%%%%%%%%%%%%%%%%%%%
%%%%%%%%%%%%%%%%%%%%%%%%%%%%%%%%%%%%%%%%%%%%%%
%%%%%%%%%%%%%%%%%%%%%%%%%%%%%%%%%%%%%%%%%%%%%%
%
\section{Sympathetic Cooperative Driving Framework}
\label{sec:solution}
In their recent work, Silver\etal explained how artificial intelligence agents can learn complex tasks through experience and maximizing a generic reward function, rather than requiring task-specific specialized problem formulations~\cite{silver2021reward}. Inspired by this approach to solving decision-making problems, rather than breaking down our problem into \textit{learning how to drive} and \textit{learning social coordination}, we train our autonomous agents from scratch using a decentralized reward structure and expect them to master the basics of highway driving, e.g., avoiding collisions and unnecessary lane change or acceleration, while learning inter-agent coordination to eventually achieve the goal of enabling a safe and seamless merging. To reiterate on our goal, we seek a decentralized solution that enables the autonomous agents to make independent socially-desirable decisions, with no explicit coordination or sharing of their decisions and future actions. In the rest of this section, we define the action and observation space in the POSG framework of Section~\ref{sec:preliminaries} and introduce the notions of sympathy and cooperation that are essential for structuring the reward function.
\subsection{Action and Observation Spaces}
\label{sec:spaces}
We employ a numeric representation for an agent's observation that embeds the kinematics of the neighboring vehicles. Additionally, we integrate the history of vehicles' last $h$ meta-actions to extract temporal information and their past trajectories. An ego vehicle $I_i \in \mathcal{I}$ observes a set of HVs and AVs in its perception range. The \textit{Kinematic} observation includes the relative Frenet coordinates of the closest $|\widetilde{\mathcal{I}} \cup \widetilde{\mathcal{V}}|+1$ vehicles in addition to the absolute Frenet coordinates of the ego vehicle. Formally, agent $I_i$ receives a local observation $\Tilde{\textbf{o}}_{i} \in \widetilde{\mathcal{O}}_i$,
\begin{equation} \label{equ:kinematicobservation}
\Tilde{\textbf{o}}_{i} = \big[ o_i, o_m, o_{i+1}, ..., o_{i+|\widetilde{\mathcal{I}} \cup \widetilde{\mathcal{V}}|} \big]^\top
\end{equation}

Each row of the local observation matrix $\textbf{o}^{(i)}$ is defined as,
\begin{equation} \label{equ:neighborvehiclestate}
o_j = \Big[ p_j, l_j, d_j, \mathrm{d}l_j/\mathrm{d}t, \mathrm{d}d_j/\mathrm{d}t, \cos \rho_j, \sin \rho_j, \lambda_j, \overline{\mathrm{H}}^{\mathcal{A}}_j \Big]
\end{equation}
in which, $l_j$ and $d_j$ are the longitudinal and lateral Frenet coordinates of the $j$th vehicle, respectively. Vehicle's yaw angle is denoted by $\rho$ and the autonomy flag is $\lambda_j = 0$ if $I_j \in \widetilde{\mathcal{V}}$ and $\lambda_j = 1$ otherwise. In case that the total number of observed vehicles is smaller than the set size of the observation matrix $\textbf{o}$, the remaining rows are filled with zeros with $p_j=0$. $\overline{\mathrm{H}}^{\mathcal{A}}_j$ is the unrolled numeric representation of the action history array $\textbf{H}^{\mathcal{A}}_j$ that contains the last $h$ meta-actions taken by $I_j$ and is defined as,
\begin{equation} \label{equ:actionhistoryarray}
\textbf{H}^{\mathcal{A}}_j(t) = \big[ a_j(t-1), ...,a_j(t-h) \big]
\end{equation}

Our interest is in maneuver-level decision-making for autonomous vehicles. Thus, we define the action space $\mathcal{A}$ as the set of abstract meta-actions $\mathcal{A}_i = [\texttt{Lane Left}$, $\texttt{Idle}$, $\texttt{Lane Right}$, $\texttt{Accelerate}$, $\texttt{Decelerate}]^\top$. These meta-actions are then translated into admissible trajectories and low-level control signals that eventually govern the movement of the vehicle. The implementation details of how meta-actions render into steering and acceleration signals are discussed in Section~\ref{sec:implementation}. Additionally, the discrete meta-actions defined above must be translated into numeric values in Eq.~\eqref{equ:actionhistoryarray}. We experiment with three encodings and choose the one that leads to best performance after training:
\begin{itemize}
    \item [-] \textit{Binary}: A one-hot encoding with 5 bits for $a_i \in \mathcal{A}_i$.
    \item [-] \textit{Discrete}: An integer in $(0, 5]$ for $a_i \in \mathcal{A}_i$.
    \item [-] \textit{Frenet}: Two integers in $[-1, 1]$ for lateral and longitudinal actions.
\end{itemize}
\subsection{Disentangling Sympathy and Cooperation}
\label{sec:sympcoop}
Inter-agent relations in our mixed-autonomy problem can be broken down into the interactions among autonomous agents, i.e., AV-AV interactions, as well as between autonomous agents and human drivers, i.e., human-AI interactions. Decoupling the two enables us to systematically study the interactions between human drivers with ambiguous SVO and our autonomous agents. We refer to an autonomous agent's altruism toward a human as \textit{sympathy} and define \textit{cooperation} as the altruistic behavior among autonomous agents. Our rationale for decoupling the components of altruism is that they differ in nature. As an instance, sympathy may not be reciprocal as humans are heterogeneous in their SVO but cooperation among autonomous agents is essentially homogeneous, assuming that they hold the same SVO. We investigate each component of altruism separately to better understand the emerging behaviors and the mechanics of inducing altruism in autonomous agents. Following this definition, we can rewrite Eq.~\eqref{equ:svodefinition} as,
\begin{equation} \label{equ:sympcoop}
\begin{aligned}
R_i  ={} & r_i \cos \phi_i +(\sin \theta_i R_i^{\mathrm{AV}} + \cos \theta_i R_i^{\mathrm{HV}}) \sin \phi_i \\
 = &  \underbrace{r_i \cos \phi_i}_{\mathrm{egoistic \,\, term}} + \\
& \underbrace{\sin \theta_i \sin \phi_i R_i^{\mathrm{AV}}}_{\mathrm{cooperation \,\, term}} + \underbrace{\cos \theta_i \sin \phi_i R_i^{\mathrm{HV}}}_{\mathrm{sympathy \,\, term}}
\end{aligned}
\end{equation}
where $\theta$ is the sympathy angular phase determining the cooperation-to-sympathy ratio. Parameters $R_i^{\mathrm{AV}}$ and $R_i^{\mathrm{HV}}$ denote the total utility of other autonomous and human-driven vehicles, respectively, as perceived from the $i$th agent's perspective. We expand on this topic in Section~\ref{sec:reward} where we introduce the distributed reward structure.
\subsection{Decentralized Reward Structure}
\label{sec:reward}
Following the notions of sympathy and cooperation and the notation of Eq.~\eqref{equ:sympcoop} we decompose the decentralized reward received by agent $I_i \in \mathcal{I}$ as,
\begin{equation} \label{equ:decentralizedreward}
\begin{aligned}
R_i(s_i, a_i) ={} & R^{\mathrm{E}}+R^{\mathrm{C}}+R^{\mathrm{S}}
\\={} &  r_i(s_i, a_i) \cos \phi_i \\
& + \sin \theta_i \sin \phi_i \sum_j \Big( r^{\mathrm{AV}}_{i, j} (\Tilde{\textbf{o}}_i) + r_j^M (\Tilde{\textbf{o}}_i) \Big) \\
& + \cos \theta_i \sin \phi_i \sum_k \Big( r^{\mathrm{HV}}_{i, k} (\Tilde{\textbf{o}}_i) + r_j^M (\Tilde{\textbf{o}}_i) \Big)
\end{aligned}
\end{equation}
in which $j \in \widetilde{\mathcal{I}} \setminus \{I_i\}$, $k \in (\widetilde{\mathcal{V}} \cup \{M\}) \setminus (\mathcal{I} \cap \{M\})$. The $r_i$ term denotes the ego vehicle's driving performance derived from metrics such as distance traveled, average speed, and a negative cost for changes in acceleration to promote a smooth and efficient movement by the vehicle. The cooperative reward term, $r^{\mathrm{AV}}_{i, j}$ accounts for the utility of the ego's allies. It is important to note that ego vehicle only requires the observation $\Tilde{\textbf{o}}_i$ to compute $R^{\mathrm{C}}$ and not any explicit coordination or knowledge of the actions of the other agents. The sympathetic reward term, $r^{\mathrm{HV}}_{i, k}$ is defined as
\begin{equation} \label{equ:symreward}
r^{\mathrm{HV}}_{i, k} = \sum_k \frac{1}{\eta d_{i,k}^\psi}u_k,
\end{equation}
where $u_k$ denotes an HV's utility, e.g., its speed, $d_{i,k}$ is the distance between the observer autonomous agent and the $k$th HV, and $\eta$ and $\psi$ are dimensionless coefficients. Moreover, the sparse scenario-specific \textit{mission reward} term $r^{\mathrm{M}}_e$ in the case of our driving scenario is representing the success or failure of the merging maneuver,
\begin{equation}
\label{equ:missionreward}
r^{\mathrm{M}}_e = 
\begin{cases}
1/2,              & \text{if $I_e \equiv M$ and merge is successful} \\
0,                  & \text{o.w.}
\end{cases}
\end{equation}

\subsection{Deep MARL for Sympathetic and Cooperative Driving}
\label{sec:DMARL}
Two cascade multi-layer perceptron (MLP) networks are utilized as the feature extractor network (FEN) and the function approximator network (FAN), each with two layers of size 256 and 128 neurons, respectively, and rectified linear unit (ReLU) non-linearities. As introduced in Section~\ref{sec:spaces}, the temporal information in a vehicle's observations is captured through integrating the history of the past actions in the observations and the feature extractor network must be able to efficiently extract meaningful patterns from this information. Both networks are trained end-to-end to enforce the feature extractor network to extract the most vital information that is required for estimating the state-action value function. The policy is trained offline and deployed into all agents to be executed in a distributed and online fashion, meaning that each agent makes independent decisions based on its observation but they all follow the same stochastic policy.

As we elaborated in Section~\ref{sec:preliminaries}, the non-stationarity of the environment is a major problem in concurrent training of multiple RL agents. We employ a semi-sequential training and policy dissemination algorithm to cope with this challenge and stabilize the training process. Algorithm~\ref{alg:MARL_algorithm} summarizes our overall methodology which is done in two stages. First, an experience replay buffer (ERB) is filled with data from simulation episodes and then, random samples drawn from this buffer is used for updating the weights of both FEN and FAN networks. For simplicity we refer to the set of all weights for both neural networks as $\textbf{w}$. We use a novel method for scoring the entries in ERB and drawing them with a probability proportional to that score.

ERB is highly skewed due to the nature of our highway merging scenario. To elaborate, each episode can be morphologically broken down into two parts, straight driving on the highway and the merging point. The former mostly provides information and training samples that are useful for learning the basics of driving and the latter contains the important information regarding the inter-agent coordination and altruistic behavior, which is of our interest. Only a few time steps of each episode contain the merging point and the rest is mostly related to highway cruising. To balance the training data drawn from the experience replay, we randomly draw samples with a probability $p_{\mathrm{ERB}}$ proportional to their spatial distance from the merging point. This method showed better performance when compared to the most common method of prioritizing the experience replay based on a sample's last resulted reward.

After drawing a training sample from ERB, the agent $I_i \in \mathcal{I}$ performs $k_{\mathrm{diss}}$ iterations of training while the weights $\textbf{w}^-_j$ of all other agents $I_j (j \neq i)$ is frozen. The updated weights $\textbf{w}^+_i$ are then disseminated to the other agents to update their policy. This process is then repeated for all agents until convergence. Doing so enables us to stabilize the training and train all agents concurrently. The key idea is applying incremental updates and keeping the environment stationary in-between the updates so that the optimizer achieves convergence. This semi-sequential algorithm is illustrated in Figure~\ref{fig:policydissemination} and Algorithm~\ref{alg:MARL_algorithm}.
\begin{algorithm}[t]
    \caption{\small{Semi-sequential multi-agent Q-learning}} 
    \label{alg:MARL_algorithm}
    \begin{algorithmic}
        \STATE Initialize experience replay buffer ($\mathrm{ERB}$) of size $N_{\mathrm{buff}}$
        \FOR{$\mathrm{Episode}=1$ to $N_{\mathrm{episode}}$}
            \STATE  Initialize episode with $l_M(t_0)$ and $v_M(t_0)$
            \FOR{$t=1$ to $T_{\mathrm{episode}}$}
                \STATE  Fill $\mathrm{ERB}$ with the tuples $([\textbf{o}_i], [a_i], [\textbf{o}'_i], [R_i])$
                \STATE Calculate the relevance factor $p_{\mathrm{ERB}}$ for each entry in $\mathrm{ERB}$
              \ENDFOR
        \ENDFOR
        \STATE  Initialize $\Tilde{Q}(s,a;\textbf{w})$ with random weights $\textbf{w}^-$
        \STATE  Initialize target network $\hat{\textbf{w}}$ with weights $\hat{\textbf{w}}=\textbf{w}^-$
        \FOR{$\mathrm{Frame} = 1$ to $N_{\mathrm{episode}} \times T_{\mathrm{episode}}$}
        \STATE  $c_{\mathrm{target}}=0$
            \FOR{$I_i$ in $\mathcal{I}$}
                \STATE  Freeze the weights $\textbf{w}^-$ for $I_j$ where $j \neq i$
                \FOR{$k=1$ to $k_{\mathrm{diss}}$}
                    \STATE Calculate the spatial distance
                    \STATE Draw a sample from ERB based on $p_{\mathrm{ERB}}$ values
                    \STATE $\textbf{w}^+ \leftarrow \textbf{w} + \alpha \hat{\nabla}_\textbf{w} \mathcal{L}(\textbf{w})$
                    \STATE  $c_{\mathrm{target}}$++
                    \IF{$c_{\mathrm{target}}$ == $n_{\mathrm{target}}$}
                        \STATE  $\hat{\textbf{w}} \leftarrow \textbf{w}^+$
                    \ENDIF
                \ENDFOR
                \STATE $\textbf{w}^- = \textbf{w}^+$ for all $I_i \in \mathcal{I}$
            \ENDFOR
        \ENDFOR
    \end{algorithmic}
\end{algorithm}
%

%
%%%%%%%%%%%%%%%%%%%%%%%%%%%%%%%%%%%%%%%%%%%%%%
%%%%%%%%%%%%%%%%%%%%%%%%%%%%%%%%%%%%%%%%%%%%%%
%%%%%%%%%%%%%%%%%%%%%%%%%%%%%%%%%%%%%%%%%%%%%%
%
\section{Implementation Details}
\label{sec:implementation}
We start this section with the 2D micro-traffic simulator we employed to generate simulation episodes and formulate the human driver model that imitates the behavior of a HV in mixed-autonomy environments. Practical details of training and validation are discussed before presenting our results in the next section.

\subsection{Driving Simulator}
\label{sec:drivingsimulator}
We modified an OpenAI Gym environment~\cite{leurent2019approximate} to enable multi-agent training and distributed execution in a mixed-autonomy highway merging scenario. The meta-actions determined by the stochastic policy are translated to low-level steering and acceleration control signals through a closed-loop proportional–integral–derivative (PID) controller. Motion of the vehicles is then governed by a Kinematic Bicycle Model that determines the vehicles' yaw rate and acceleration. As a common practice in robotics, road segments and the motion of the agents are expressed in Frenet-Serret coordinates and broken into lateral and longitudinal movements.

In order to ensure learning generalizable policies rather than memorizing a sequence of actions by the function approximator network, the initial state of each simulation episode is randomized. This episode initialization is particularly critical as the resulting initial states must be still meaningful and valid for our desired conflictive highway merging scenario. Trivial episodes where the merging vehicle can easily merge into the highway regardless of the AVs' actions or the episodes where the AVs' do not have an opportunity to enable safe merging, not only do not add valuable information to the training process but also can lead into misleading measures. The initial longitude and speed of the cruising vehicles are uniformly randomized and the initial longitude $l_{M}(t_0)$ and speed $v_{M}(t_0)$ of the merging vehicle are drawn from a clipped-Gaussian distribution $\widetilde{\mathcal{N}}(x;\mu,\sigma,\delta)$ defined as, 
\begin{equation}
\label{equ:clippedgaussian}
\widetilde{\mathcal{N}}(x)=\mathcal{N}(x;\mu,\sigma) \Big(\mathds{1}(x-\mu+\delta) - \mathds{1} (x-\mu-\delta) \Big)
\end{equation}
where $\mathcal{N}(x;\mu,\sigma)$ denotes a Gaussian distribution and $\mathds{1}$ is the Heaviside step function. We elaborate on initializing episodes via parameters $\mu$, $\sigma$, and $\delta$ in Section~\ref{sec:ablations}.

\subsection{Human Driver Model}
\label{sec:humandrivermodel}
Lateral and longitudinal movements of HVs are mimicked by human driver models proposed by Treiber\etal and Kesting~\etal~\cite{treiber2000congested, kesting2007general}. The lateral actions of HVs, i.e., the decision to perform a lane change, follow the Minimizing Overall Braking Induced by Lane changes (MOBIL) strategy~\cite{kesting2007general}. MOBIL model allows a lane change only if the resulting acceleration $\mathrm{acc}^{}_n>-b_{\mathrm{safe}}$ meets the safety criterion, and the incentive criterion is also satisfied,
\begin{equation}
\label{equ:mobilcondition}
\mathrm{acc}'_e-\mathrm{acc}^{}_e+\sin \phi_{e} \Big( (\mathrm{acc}'_n-\mathrm{acc}^{}_n)  + (\mathrm{acc}'_o-\mathrm{acc}^{}_o) \Big) > \mathrm{acc}^{}_{\mathrm{th}}
\end{equation}
with $\mathrm{acc}^{}_{e}$, $\mathrm{acc}^{}_{n}$, and $\mathrm{acc}^{}_{o}$ being the acceleration of the ego HV, the following vehicle in the target lane, and the following vehicle in the current lane, respectively, and $\mathrm{acc}'_{e}$, $\mathrm{acc}'_{n}$, and $\mathrm{acc}'_{o}$ are the corresponding accelerations assuming the ego HV has performed the lane change. $\mathrm{acc}_{\mathrm{th}}$ is the threshold that determines if the ego HV shall performs the lane change. HV's SVO angle $\phi_e$ is also referred to as the politeness factor in the literature and is extracted from the empirical probability distribution illustrated in Figure~\ref{fig:svoring}.

The longitudinal acceleration of HVs follows the Intelligent Driver Model (IDM)~\cite{treiber2000congested}. The longitudinal Frenet acceleration of a HV, $\Ddot{l}_{\mathrm{IDM}}$, is determined by
\begin{equation}
\label{equ:idm1}
\Ddot{l}_{\mathrm{IDM}}=\mathrm{acc}_\mathrm{max}\bigg( 1- \Big( \frac{\dot{l}}{v_{\mathrm{set}}} \Big)^4 - \Big( \frac{d^*(\dot{l}, \Delta \dot{l})}{d} \Big)^2 \bigg)
\end{equation}
where $\dot{l}$ denotes the longitudinal Frenet speed of the HV, and the desired Frenet distance to the leading vehicle is controlled by $d^*$, defined as,
\begin{equation}
\label{equ:idm2}
d^*(\dot{l}, \Delta \dot{l}) = d_0 + \dot{l} T_\mathrm{set} +  \frac{\dot{l} \Delta \dot{l}}{ 2\sqrt{\mathrm{acc}_{\mathrm{max}}.\mathrm{acc}_{\mathrm{des}}}}
\end{equation}
in which $\Delta \dot{l}$ is the approach rate, and the model parameters $v_{\mathrm{set}}$, $T_{\mathrm{set}}$, $d_0$, $\mathrm{acc}_{\mathrm{max}}$, and $\mathrm{acc}_{\mathrm{des}}$ are set speed, set time gap, minimum gap distance, maximum acceleration, and the desired acceleration, respectively. Additionally, the acceleration of the vehicle is a random variable defined as,
\begin{equation}
\label{equ:idm3}
\Ddot{l} = \Ddot{l}_{\mathrm{IDM}}+\frac{\sigma_{\mathrm{vel}}}{\Delta t}\mathcal{N}(0,1)
\end{equation}
with $\mathcal{N}(0,1)$ being a standard Gaussian random variable and $\sigma_{\mathrm{vel}}$ is the standard deviation of the velocity noise at the time step $\Delta t$ of the simulation.

\begin{figure}[t]
  \centering
  \includegraphics[width=.48\textwidth]{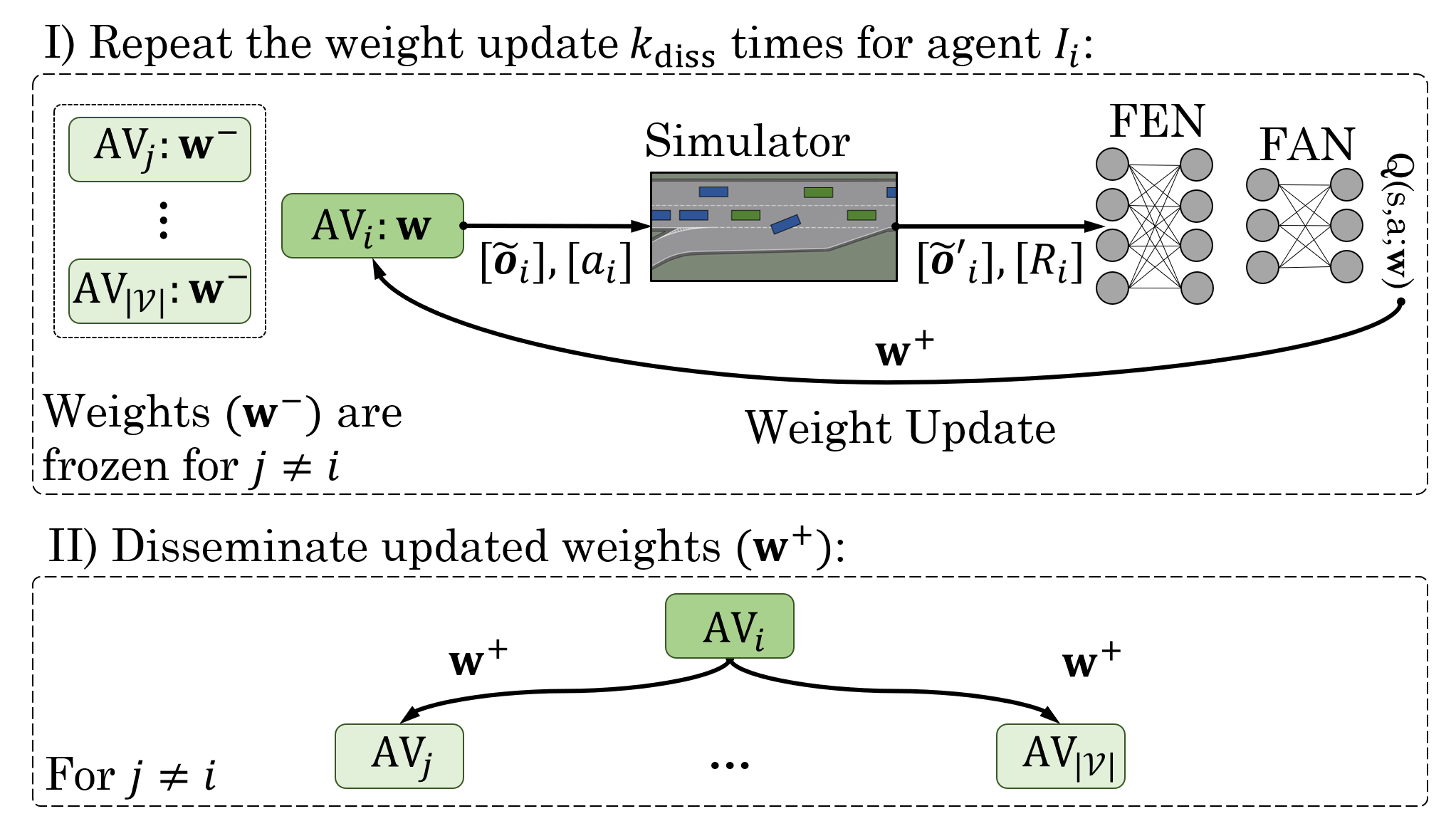}
  \caption{\small{Multi-agent training and policy dissemination process.}}
  \label{fig:policydissemination}
\end{figure}

\subsection{Training and Validation}
The autonomous agents are trained using the semi-sequential multi-agent Q-learning algorithm that we introduced in Figure~\ref{fig:policydissemination} and Algorithm~\ref{alg:MARL_algorithm} for 15,000 episodes that are generated by the procedure discussed in Section~\ref{sec:drivingsimulator}. Training process is repeated and compared across multiple runs to assure the stability of training and that it converges to the similar policies every time. The trained policies are then evaluated for 2,000 randomized novel test episodes to gauge their efficacy. Test episodes are intentionally generated with different and broader initialization range than the training episodes to demonstrate that agents actually are able to learn generalizable policies and not only memorize sequences of actions.

%
%%%%%%%%%%%%%%%%%%%%%%%%%%%%%%%%%%%%%%%%%%%%%%
%%%%%%%%%%%%%%%%%%%%%%%%%%%%%%%%%%%%%%%%%%%%%%
%%%%%%%%%%%%%%%%%%%%%%%%%%%%%%%%%%%%%%%%%%%%%%
%
\section{Experimental Results}
\label{sec:experiments}
We break down the research questions of our interest into experimental hypotheses and investigate them through our experiments and ablation studies in this section.

\begin{figure}[t]
  \centering
  \includegraphics[width=.5\textwidth]{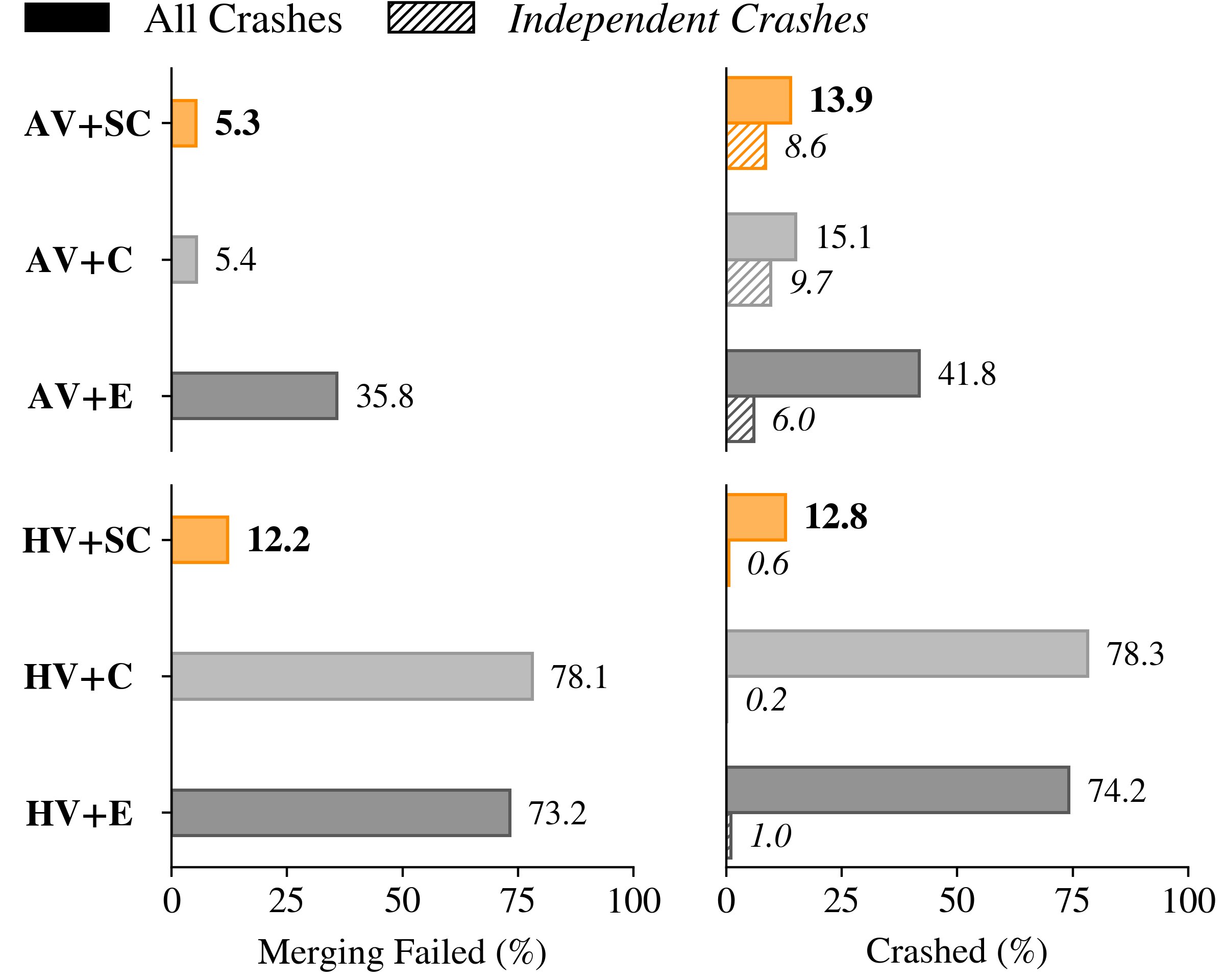}
  \caption{\small{Impact of \textit{sympathy} and \textit{cooperation} elements in traffic safety and success of the merging maneuver for both $M\in\mathcal{I}$ or $M\in\mathcal{V}$. Hatched bars show the number of independent crashes that do not involve the mission vehicle.}}  
  \label{fig:fig_A}
\end{figure}

\subsection{Manipulated Variables}
\label{sec:manipulatedvariables}
The two key variables in Eq.~\eqref{equ:decentralizedreward} are $\phi$ and $\theta$ that determine the level of altruism, which is the general term we use for both HVs and AVs, as well as level of sympathy, which is the term for altruism toward HVs only. Our experiments are done in 2$\times$6 settings with different values of $phi$ and $\theta$. Furthermore, we experiment with both autonomous, $M \in \mathcal{I}$, and human-driven, $M \in \mathcal{V}$, mission vehicle. Our experiment settings are:
\begin{itemize}
    \item \textbf{HV+E. } autonomous agents are egoistic ($\phi_i = 0$ for $I_i \in \mathcal{I}$), and the mission vehicle is HV ($M \in \mathcal{V}$);
    \item \textbf{HV+C. } autonomous vehicles are cooperative only ($\phi_i = \phi^*$ and $\theta_i = \pi/2$ for $I_i \in \mathcal{I}$), and the mission vehicle is HV ($M \in \mathcal{V}$);
    \item \textbf{HV+SC. } autonomous vehicles are sympathetic and cooperative ($\phi_i = \phi^*$ and $\theta_i = \pi/4$ for $I_i \in \mathcal{I}$), and the merging vehicle is HV ($M \in \mathcal{V}$);
    \item \textbf{AV+E/C/SC.} Duals of the the above cases with autonomous mission vehicle ($M \in \mathcal{I}$).
\end{itemize}

In \textbf{HV+SC} and \textbf{AV+SC} scenarios where autonomous agents have both sympathy and cooperation components, we set the sympathy angle to $\theta=\pi/4$ for the sake of fairness and to avoid imposing bias between HVs and AVs as they both carry humans or goods and neither should have a pre-assumed advantage over the other. The SVO angle $\phi$ is however tuned to reach the optimal level of altruism, we elaborate on this topic in Section~\ref{sec:results} and derive the optimal SVO angle $\phi^*$.

\vspace{-0.1cm}
\subsection{Performance Measures}
\label{sec:performancemeasures}
To gauge the impact of the aforementioned manipulated variables and other configurable parameters, 3 metrics are chosen that despite being correlated with each other, provide different insights on the efficacy of our solution. As a traffic-level metric, the average distance traveled by HVs and AVs is logged during simulation episodes. Additionally, counting the percentage of the episodes that experience a successful merge enables us to probe the overall social importance of a solution. Safety is also gauged through counting the percentage of episodes that contain at least one crash.

\vspace{-0.2cm}
\subsection{Hypotheses}
\label{sec:hypotheses}
Social and individual performance of altruistic and purely egoistic agents are compared through the 3 key hypotheses:
\begin{itemize}
    \item \textbf{H1.} \emph{While egoistic AVs fail to account for a merging HV, AVs that hold both sympathy and cooperation elements explore ways to enable safe and seamless merging. Therefore, we expect \textbf{HV+SC} to outperform \textbf{HV+E} and \textbf{HV+C} settings.}
    \item \textbf{H2.} \emph{AVs with $\phi \neq 0$ are able to implicitly learn the SVO of HVs and guide them to improve the overall performance of the group.}
    \item \textbf{H3.} \emph{There exists a social value orientation angle ${0<\phi^*<\pi/2}$ for autonomous agents that can both lessen the number of crashes and improve the number of successful merges.}
\end{itemize}

\begin{figure}[t!]
\centering
\includegraphics[width=.49\textwidth]{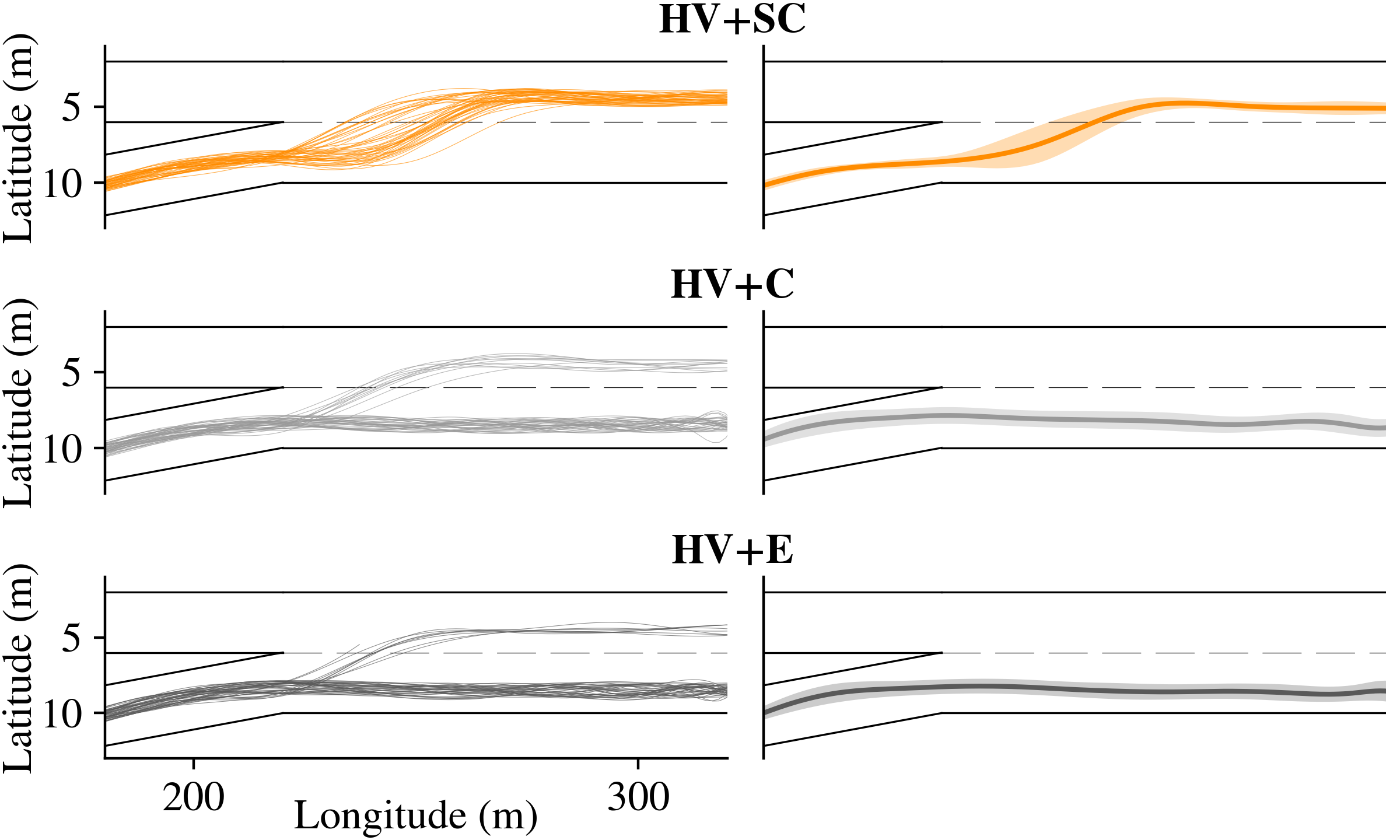}
\caption{\small{Sampled trajectories of the human-driven mission vehicle $M$ shows the efficacy of \textbf{SC} agents. Mean and standard deviation shown on the right hand side plots.}}
\label{fig:fig_B}
\end{figure}

\subsection{Analysis and Results}
\label{sec:results}

\smallskip
\noindent \textbf{Examining H1. }
The main claim of hypothesis \textbf{H1} is the superiority of sympathetic cooperative AVs in creating socially optimal results when compared to egoistic autonomous AVs. To better understand the situation, we reiterate on the driving scenario: the merging vehicle $M$, that can be either human-driven or autonomous, approaches a highway with a mixed group of HVs and AVs. $M$ requires the cruising vehicles' assistance in order to be able to merge safely. Per our fundamental assumption, we do not rely on the HVs to compromise on their own utility as their SVO is unknown. Instead, it's on the AVs to create a safe corridor for $M$ and, as we will show in Section~\ref{sec:ablations}, this goal cannot be achieved by a single AV alone and necessarily needs a cooperative action by the group of AVs.

\begin{figure*}[t!]
\centering
\includegraphics[width=.95\textwidth]{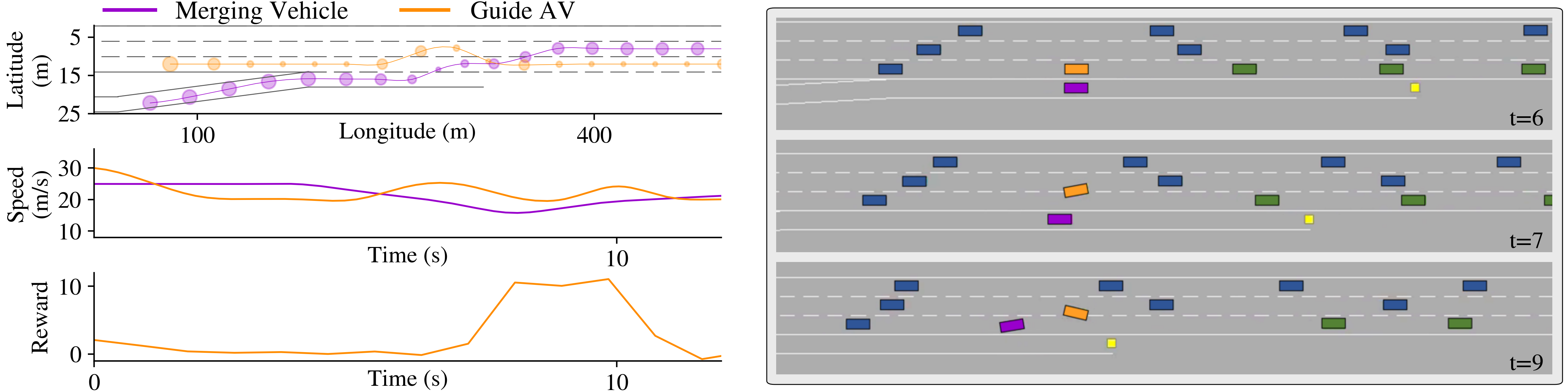}
\caption{\small{An example of AVs (\textit{green}) and HVs (\textit{blue}) in \textbf{HV+SC} setting. Successful merging requires all the AVs to work together and none of them can achieve this goal alone. We focus on the ``Guide AV" that makes the most significant impact and show how it learns to take sequences of actions to not only enable the mission vehicle to merge (by decelerating and performing a lane change to left), but also manages to make the minimal compromise on its individual utility (by another lane change to right and cruising with optimal speed). Diameter of the circles on the trajectory plot shows vehicles' speed.}}
\label{fig:fig_D_wide}
\end{figure*}

Figure~\ref{fig:fig_A} illustrates an overall comparison between the settings defined in Section~\ref{sec:manipulatedvariables}. Focusing on the cases with a human-driven merging vehicle, it is evident that in the absence of the \textit{sympathy} component in AVs, i.e., in \textbf{HV+E} and \textbf{HV+C} settings, merging fails in the majority of episodes. Failed merging leads to a crash in our simulator as vehicles cannot stop on the highway nor the merging ramp and the merging vehicle that fails to merge collides with the barrier at the end of the merging ramp. This assumption is made to make our simulations more realistic and avoid unfeasible solutions that require full-stop on the highway. Therefore, most of the crash cases shown in Figure~\ref{fig:fig_A} are due to unsuccessful merging and not the lack of basic driving skills in HVs and AVs. As an additional evidence, independent crashes that are not relevant to a failed merge are also plotted in Figure~\ref{fig:fig_A}, which confirms the fact that the vehicles hold sufficient basic skills to maneuver on a highway and avoid collisions. 

Figures~\ref{fig:fig_A}~and~\ref{fig:fig_B} clarify the positive social impact that sympathy and cooperation make in terms of reducing the total number of crashes and failed merging. However, a counter-argument against this comparison can be the fact that a rather conservative model is used to mimic HVs in our simulations and this might limit their capability in merging. To investigate this claim, we repeat the comparison with an autonomous mission vehicle that is more risk-tolerant and attempts more creative ways to merge into the highway. In the \textbf{AV+E} setting that AVs only care about their individual utility, although the results are better compared to \textbf{HV+E}, even the autonomous mission vehicle still fails to safely merge in more than 1/3 of the episodes. We conclude that our test case indeed creates a competitive and conflictive scene for the vehicles and showcases how incorporating sympathy and cooperation components in the reward structure of AVs leads to socially-desirable outcomes and improves safety and traffic flow. Figure~\ref{fig:fig_B} provides further intuition to this comparison by depicting a sampled set of mission vehicle's trajectories in different experiment settings. It is evident that un-sympathetic does not allow the mission vehicle to merge, causing its trajectory to end in the merging ramp.

\smallskip
\noindent \textbf{Examining H2. }
Figure~\ref{fig:fig_D_wide} illustrates an example of autonomous agents trained with the sympathetic cooperative reward and a higher capacity neural network architecture. Although all AVs in this scenario work together to make the merging possible, we focus on the most impactful agent which is the ``Guide AV" shown in orange color. Other AVs in this sample scenario (shown in green) compromise on their individual reward by accelerating, consuming more energy, and thus receiving less reward as defined in Section~\ref{sec:reward}. Interestingly, the Guide AV learns to first slow down and then change lane to left and open up space for $M$. After $M$ successfully merges, the Guide AV finds its lane blocked by a HV so makes another lane change to the right and follows other AVs. Figure~\ref{fig:fig_D_wide} demonstrates how AVs receive a significant reward when $M$ merges into the highway. Although the reward structure defined in Section~\ref{sec:reward} contains multiple parameters but the mission reward term $r^M$ of Eq.~\eqref{equ:missionreward} has an order of magnitude larger impact and thus is the dominating reward signal in training our autonomous agents. In other words, the trained agents learn to take sequences of actions that lead to receiving $r^M$. This learning process includes learning to avoid collisions, navigating through the traffic, and if required affecting the behavior of other HVs.

As it was emphasized before, the autonomous agents do not have access to an explicit behavior model of human drivers and instead implicitly learn this model from experience during the training episodes. Although we employ a rather conservative model of human drivers to showcase our concept, it is expected that given sufficient training data, the autonomous agents can extract models of more complex human behaviors as well. However, sensitivity of our solution to these models and the effect of human behaviors on inter-agent coordination is a topic worthy of investigation which we leave for our future work. As a relevant observation, AVs implicitly learn to predict the behavior of HVs and the fact that HVs commonly act egoistically (refer to Figure~\ref{fig:svoring}) and do not slow down for the merging vehicle. Hence, they do not rely on the HVs and instead compromise on their individual reward to enable the highway merging.

\smallskip
\noindent \textbf{Examining H3. }
\begin{figure}[t]
\centering
\includegraphics[width=.49\textwidth]{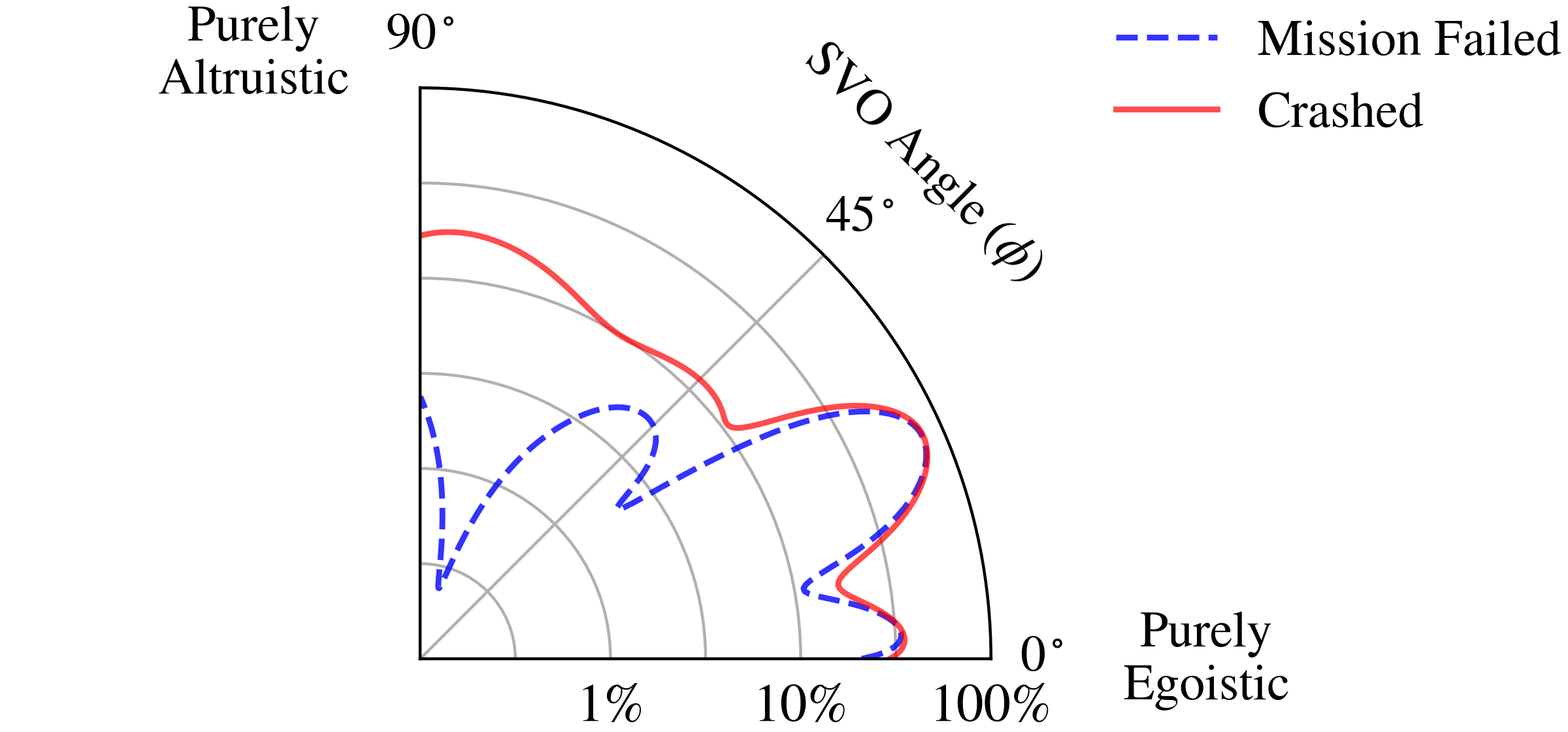}
\caption{\small{Finding the optimal SVO angular phase $\phi^*$ for AVs that results the least number of crashes and failed merges.}}
\label{fig:Fig_F}
\end{figure}
Our experimental scenarios in Section~\ref{sec:manipulatedvariables} are defined based on the optimal SVO angle $\phi^*$ of the autonomous agents. This parameter clearly has an important impact on the behavior of AVs and thus the safety and traffic-flow metrics. We trained a large set of agents with different SVO angles and tested them in our case study driving scenario. The optimal SVO angle is then defined as the angle that results the best performance metrics, i.e., least number of episodes with collisions and failed merges. We formulate this simple optimization objective as the convex combination of the two metrics,
\begin{equation}
\label{equ:optimalsvo}
\phi^*=\argmin_\phi \Big( \xi . f_{\mathrm{C}}(\phi) + \big( 1-\xi \big) . f_{\mathrm{MF}}(\phi) \Big)
\end{equation}
where $f_{\mathrm{C}}$ and $f_{\mathrm{MF}}$ are the percentage of episodes with a crash and failed mission, respectively. The hyper-parameter $\xi$ determines the importance of each performance metric and we choose it to be $\xi=0.5$ as otherwise it could bias the training process by putting more emphasis on either of the metrics. Figure~\ref{fig:Fig_F} illustrates how the two metrics change when the autonomous agents' SVO is varied from $\phi=0$ (purely egoistic) towards $\phi=\pi/2$ (purely altruistic). It is worth mentioning that neither of the two extremes seems optimal and a point between \textit{caring about others} and \textit{being selfish} leads to the most socially-desirable outcome.

\begin{table}[b]
\caption{\small{Necessity of Multi-agent Coordination: a single SC agent is not able to create socially-desirable outcomes.}}
\begin{center}
\begin{tabular*}{0.48\textwidth}{p{0.16\textwidth}   P{0.08\textwidth}  P{0.08\textwidth} P{0.08\textwidth}}
 &
Mission Failed&
Crashed&
Distance Traveled\\ 
\hline
\hline

Single-agent (\textbf{HV+1SC}) &
$74.4\%$ &
$74.5\%$ &
$268.5 m$ \\ 
Multi-agent (\textbf{HV+SC}) &
$\textbf{12.2}\%$ &
$\textbf{12.8}\%$ &
$\textbf{334.4}m$ \\ 

\hline
\end{tabular*}
\end{center}
\label{table:ablationmultiagent}
\end{table}
\begin{table}[b]
\caption{\small{Ablation study on representing agent observation $\Tilde{\textbf{o}}_i$.}}
\begin{center}
\begin{tabular*}{0.48\textwidth}{p{0.18\textwidth}   P{0.11\textwidth}  P{0.1\textwidth}}
 &
Mission Failed&
Crashed\\ 
\hline
\hline
\multicolumn{3}{l}{\textit{Adding Autonomy Flag $\lambda$}} \\ 
 \hline
Without &
$5.0\%$ &
$10.4\%$ \\ 
\textbf{With} &
$\textbf{3.8}\%$ &
$\textbf{8.9}\%$ \\ 

\hline
\hline
\multicolumn{3}{l}{\textit{Including Mission Vehicle $o_M$}} \\ 
\hline
Without &
$4.2\%$ &
$9.2\%$ \\ 
\textbf{With} &
$\textbf{1.7}\%$ &
$\textbf{8.3}\%$ \\ 
\hline
\end{tabular*}
\end{center}
\label{table:ablationsencoding}
\end{table}

\begin{figure}[t]
  \centering
  \includegraphics[width=.5\textwidth]{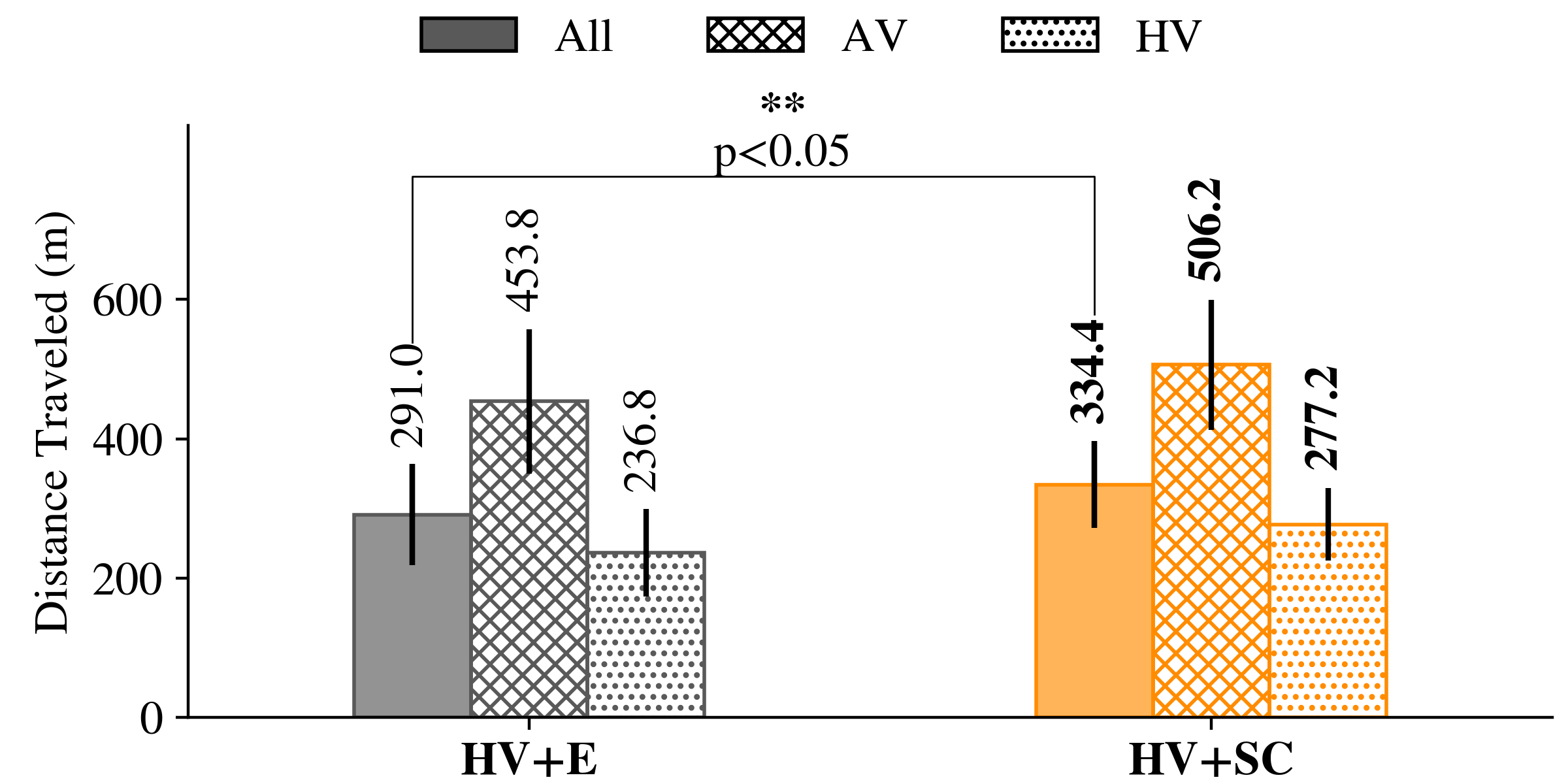}
  \caption{\small{Despite making a compromise and slowing down a group of vehicles to allow the mission vehicle to merge, \textbf{SC} agents still lead to better overall traffic flow for both AVs and HVs.}}
  \label{fig:fig_C}
\end{figure}

A fair critique to the behavior of sympathetic cooperative agents can be the fact that the Guide AV, i.e., AV3 in Figure~\ref{fig:formalscenario}, decelerates and therefore slows down the group of vehicles behind only to allow the mission vehicle to merge. In other words, the utility of a big group of vehicles is being compromised for the sake of the mission vehicle. To investigate the fairness and effectiveness of this outcome, we measure the average distance traveled by HVs and AVs. Figure~\ref{fig:fig_C} reveals how despite the fact that in the \textbf{HV+SC} setting a group of vehicles need to slow down to open up space for the mission vehicle, eventually both HVs and AVs manage to travel more distance when compared to a similar setup with egoistic agents (\textbf{HV+E}). It should be noted that the effect of Guide AV's deceleration gradually propagates through the platoon of vehicles in behind and only affects a limited group of vehicles as the traffic in the platoon is not rigid and can contract and expand.

\subsection{Ablation Studies}
\label{sec:ablations}

\smallskip
\noindent \textbf{Necessity of Multi-agent Coordination. }
Consider the highway merge scenario of Figure~\ref{fig:formalscenario}. Our claim is that all AVs require to work together to enable a safe and seamless merging and none of them can achieve this goal if the others do not cooperate. As elaborated in Section~\ref{sec:drivingsimulator}, we particularly design our scenarios to gauge the effectiveness of altruistic agents and inter-agent coordination. To complement our results in Figure~\ref{fig:fig_A} that back the hypothesis \textbf{H1}, we conducted an ablation study in the driving scenario of  Figure~\ref{fig:formalscenario} with the difference that only $\mathrm{AV3}$ is sympathetic cooperative and label this scenario as \textbf{HV+1SC}. Table~\ref{table:ablationmultiagent} demonstrate the necessity of multi-agent coordination and the fact that a single sympathetic cooperative AV, i.e., the Guide AV, is not able to achieve the mission of safe and seamless merging without help from the other AVs.

\begin{figure}[t]
\centering
\includegraphics[width=.49\textwidth]{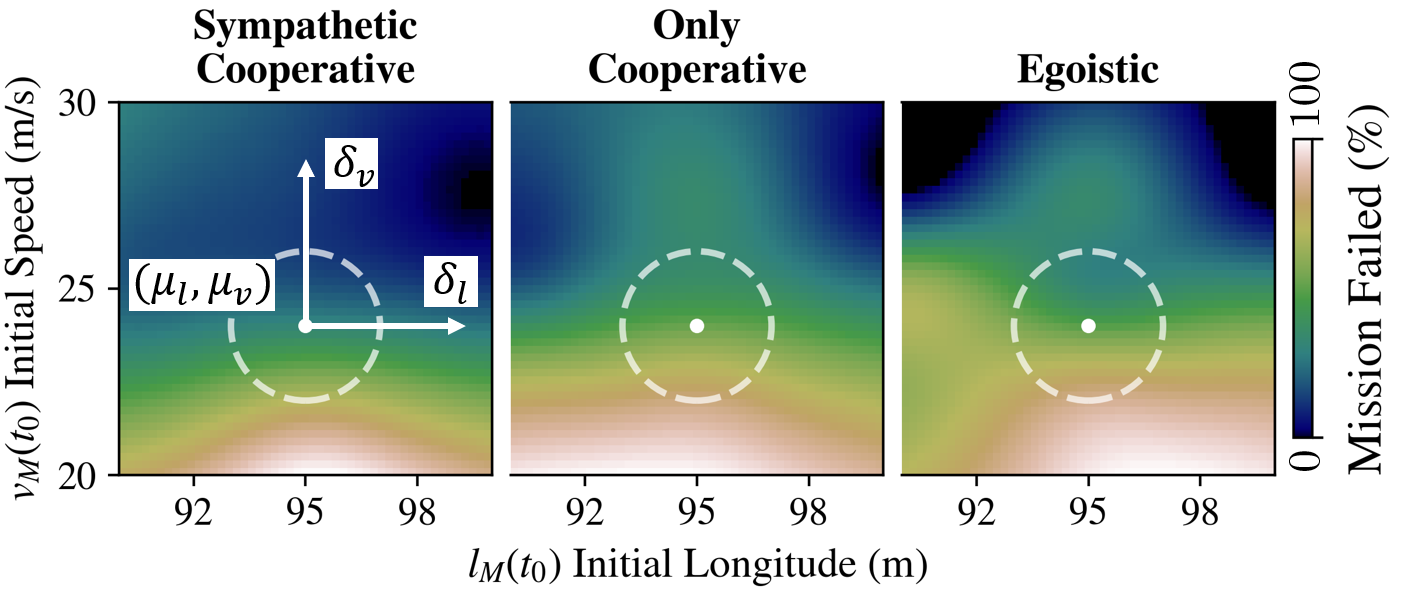}
\caption{\small{Training episodes should not be trivial nor should they benefit a specific setting (\textbf{E}, \textbf{C}, \textbf{SC}).}}
\label{fig:initialization}
\end{figure}

\smallskip
\noindent \textbf{Designing Non-trivial and Fair Scenarios. }
Our method for initializing simulation episodes is described in Eq.~\eqref{equ:clippedgaussian}. Parameters $\mu$ and $\delta$ determine the range of the allowed values for the merging vehicle's initial longitude and speed. Trivial episodes that are too easy, i.e., always lead to successful merging, or too challenging, i.e., never result in a successful merge, can steer the training process into the wrong direction and must be avoided when initializing the episodes. Furthermore, the initial state of an episode can benefit different agents with various SVOs and thus, one may argue that the superior performance of sympathetic cooperative agents as observed in Figures~\ref{fig:fig_A}~and~\ref{fig:fig_B} is an artifact of the episode's initialization. We draw the initial values from a region that does not favor either of the social preferences. Two sets of parameters $(\mu_l, \delta_l, \sigma_l = 2 \delta_l)$ and $(\mu_v, \delta_v, \sigma_v = 2\delta_v)$ are chosen for the initial longitude $l_{M}(t_0)$ and initial speed $v_{M}(t_0)$ of the merging vehicle, as listed in Table~\ref{table:params}. Figure~\ref{fig:initialization} illustrates the intuition behind choosing these values.

\begin{figure}[t]
\centering
\includegraphics[width=.49\textwidth]{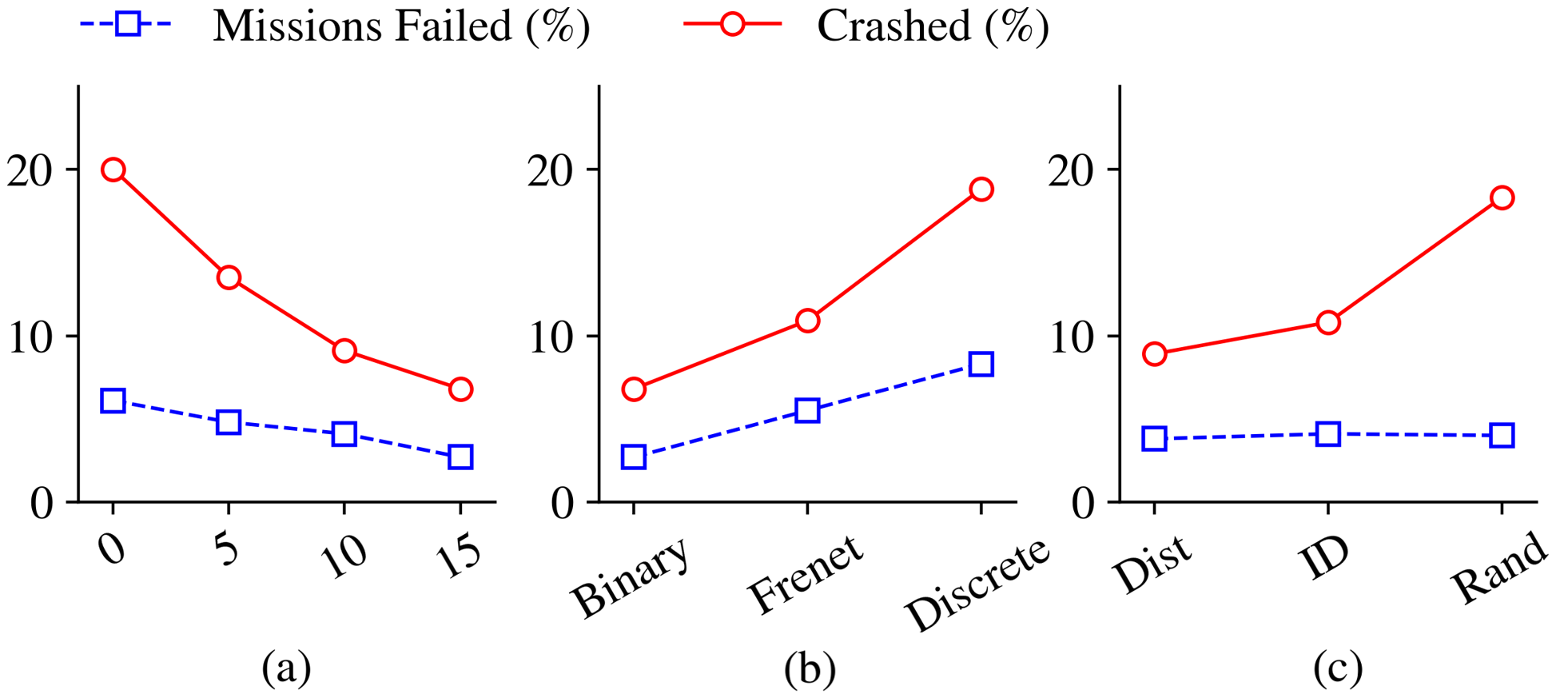}
\caption{\small{\textit{(a)} length of action history $h$, \textit{(b)} embedding type (Section~\ref{sec:spaces}), \textit{(c)} Sorting rows of Eq.~\eqref{equ:kinematicobservation} using longitudinal distance.}}
\label{fig:ablation}
\end{figure}

\smallskip
\noindent \textbf{Observation-space Representation. }
We discussed the details of how information is embedded into an agent's observation in Section~\ref{sec:spaces}. Here we justify the design choices and show their positive impact on the performance. Table~\ref{table:ablationsencoding} shows the impact of including $o_m$ in Eq.~\eqref{equ:kinematicobservation} as well as the autonomy flag $\lambda$ of Eq.~\eqref{equ:neighborvehiclestate}. Figure~\ref{fig:ablation} summarizes the effect of integrating $\overline{\mathrm{H}}^{\mathcal{A}}_j$ in Eq.~\eqref{equ:neighborvehiclestate}, the history horizon $h$, and the type of the action encoding. We also experimented with sorting the rows of $\textbf{o}^{(i)}$ in Eq.~\eqref{equ:kinematicobservation} based on vehicle ID and vehicles' longitude, as shown in Figure~\ref{fig:ablation}.

\begin{table}[b]
\caption{\small{Training and simulation hyper-parameters.}}
\begin{center}
\begin{tabular}{c c | c c}
Parameter &
Value &
Parameter &
Value\\ 
\hline
\hline
$N_{\mathrm{episode}}$ &
$10,000$ &
$\mu_l$ & %Initial position of $\mathcal{M}$ ($\mu_l$)
$95 m$ \\ 
Batch size &
$32$ &
$\delta_l$ & %$\sigma$ of $\mathcal{M}$' initial position ($\sigma_l$)
$2 m$ \\ 
$N_{\mathrm{buffer}}$ &
$100,000$ &
$\mu_v$ & %Initial velocity of $\mathcal{M}$ ($\mu_v$)
$24 m/s$ \\ 
$\alpha_0$ &
$0.0005$ &
$\delta_v$ & %$\sigma$ of $\mathcal{M}$' initial velocity ($\sigma_v$)
$2 m/s$ \\ 
$n_{\mathrm{target}}$ &
$200$ &
$v_{\mathrm{set}}$ & %Desired speed 
$25m/s$ \\
Initial exploration $\epsilon_0$ &
$1.0$  &
$T_{\mathrm{set}}$ &  %Desired time gap ($T_{\mathrm{set}}$)
$0.5 s$ \\ 
Final exploration $\epsilon_f$ &
$0.1$ &
$d_0$ & %Minimum gap distance ($d_0$)
$1 m$ \\ 
$\epsilon-$decay &
$\mathrm{Linear}$ &
$\mathrm{acc}_{\mathrm{max}}$ & %Maximum acceleration ( $\mathrm{acc}_{\mathrm{max}}$) 
$3 m/s^2$ \\ 
Optimizer &
$\mathrm{ADAM}$ &
$\mathrm{acc}_{\mathrm{des}}$   & %Desired deceleration ($\mathrm{acc}_{\mathrm{des}}$)
$-5 m/s^2$  \\ 
$\gamma$&
$0.95$  &
$\mathrm{acc}^{}_{\mathrm{th}}$ & %Politeness factor ($\sin \phi_{e}$)
$0.2 m/s^2$ \\ 
$|\mathcal{V}|$ &
$20$  &
$h$ & 
$10$ \\ 
$|\mathcal{I}|$ &
$4$ &
$\xi$ & %Maximum safe deceleration ($b_safe$)
$0.5$\\ 
$T_{\mathrm{episode}}$ &
$18s$ &
$k_{\mathrm{diss}}$ & %Maximum safe deceleration ($b_safe$)
$4$ \\ 
 
\hline
\hline
\end{tabular}
\end{center}
\label{table:params}
\end{table}
%

%
%%%%%%%%%%%%%%%%%%%%%%%%%%%%%%%%%%%%%%%%%%%%%%
%%%%%%%%%%%%%%%%%%%%%%%%%%%%%%%%%%%%%%%%%%%%%%
%%%%%%%%%%%%%%%%%%%%%%%%%%%%%%%%%%%%%%%%%%%%%%
%   
\section{Concluding Remarks}
\smallskip
\noindent \textbf{Summary. }
Autonomous vehicles need to learn to co-exist with human-driven vehicles on the same road infrastructure. Deploying egoistic AVs that solely account for their individual interests on the road leads to sub-optimal and non-desirable social outcomes. In contrast, we compute the optimal SVO angle that optimizes the traffic metrics and demonstrate how altruistic AVs with the corresponding SVO can be trained to optimize a decentralized social utility that improves traffic flow, safety, and efficiency. We propose practical solutions to mitigate the non-stationarity problem in simultaneous multi-agent training and implicitly learn the behavior of human drivers from experience. Our experiments reveal that altruistic AVs are able to form alliances and affect the behavior of HVs in order to create socially-desirable outcomes that benefit the group of the vehicles.

\noindent \textbf{Limitations and Future Work. }
While this paper captures the fundamentals of social coordination and altruism in autonomous driving, many tangential aspects of the problem can be further studied. For example, we employed a conservative and limited model of human drivers. Although we expect our solution to be effective with other human behavior models as well, it is important to study its performance under different human behaviors. Also, the impact of communication imperfections and packet drops on the inter-agent coordination can be further investigated using more complex communication models than those presented in this work. On the implementation side, more advanced neural architectures such as convolutional and recurrent networks can be leveraged to capture spatial and temporal information more effectively, a direction that we plan to explore in our future work. 
%
%%%%%%%%%%%%%%%%%%%%%%%%%%%%%%%%%%%%%%%%%%%%%%%%%%%%%%%%%%%%%%%%%%%%%%%%%%%%
%%%%%%%%%%%%%%%%%%%%%%%%%%%%%%%%%%%%%%%%%%%%%%%%%%%%%%%%%%%%%%%%%%%%%%%%%%%%
%%%%%%%%%%%%%%%%%%%%%%%%%%%%%%%%%%%%%%%%%%%%%%%%%%%%%%%%%%%%%%%%%%%%%%%%%%%%
%%%%%%%%%%%%%%%%%%%%%%%%%%%%%%%%%%%%%%%%%%%%%%%%%%%%%%%%%%%%%%%%%%%%%%%%%%%%
% Can use something like this to put references on a page
% by themselves when using endfloat and the captionsoff option.
\ifCLASSOPTIONcaptionsoff
  \newpage
\fi

\bibliographystyle{IEEEtran}
\bibliography{IEEEbibs}

\vskip -2.5\baselineskip plus -1fil
%%%%%%%%%%%%%%%%%%%%%%%%%%%%%%%%%%%%%%%%%%%
\begin{IEEEbiography}[{\includegraphics[width=1in,height=1.25in,clip,keepaspectratio]{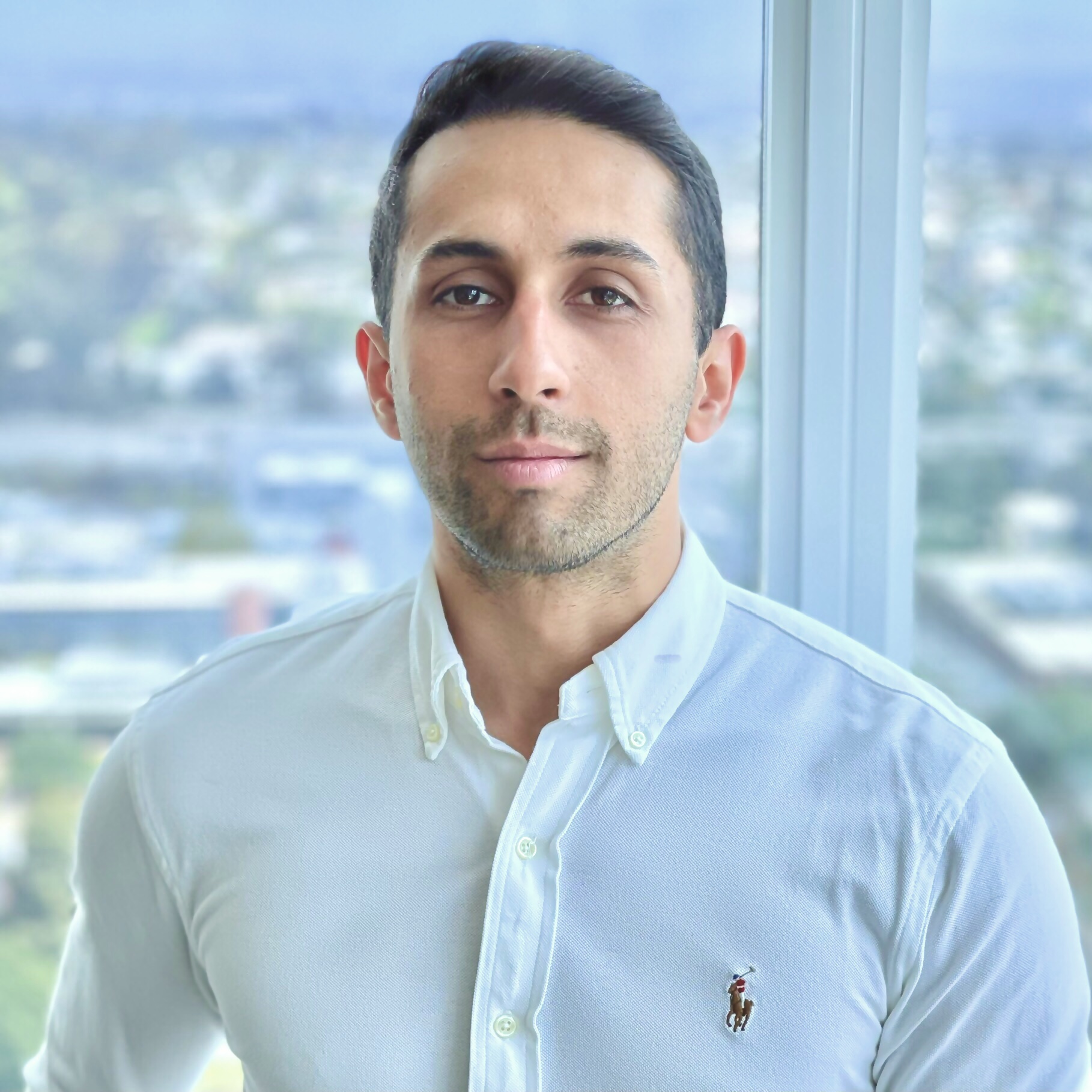}}]{Behrad Toghi}
is a Ph.D. candidate at the University of Central Florida. He received the B.Sc. degree in electrical engineering from Sharif University of Technology in 2016 and has worked as a research intern at Mercedes-Benz R\&D North America and Ford Motor Company R\&D between 2018 and 2020. His work is in the intersection of artificial intelligence and cooperative networked systems with a focus on autonomous driving.
\end{IEEEbiography}
%%%%%%%%%%%%%%%%%%%%%%%%%%%%%%%%%%%%%%%%%%%
\vskip -2.5\baselineskip plus -1fil
%%%%%%%%%%%%%%%%%%%%%%%%%%%%%%%%%%%%%%%%%%%
\begin{IEEEbiography}[{\includegraphics[width=1in,height=1.25in,clip,keepaspectratio]{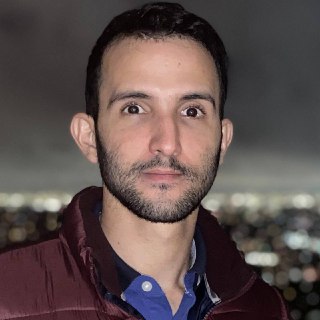}}]{Rodolfo Valiente}
is a Ph.D. candidate in Computer Engineering at the University of Central Florida. His research interests include connected autonomous vehicles (CAVs), reinforcement learning, computer vision, and deep learning with a focus on the autonomous driving problem. He received a M.Sc. degree from the University of Sao Paulo (USP) in 2017 and his B.Sc. degree from the Technological University Jose Antonio Echeverria in 2014.
\end{IEEEbiography}
%%%%%%%%%%%%%%%%%%%%%%%%%%%%%%%%%%%%%%%%%%%
\vskip -2.5\baselineskip plus -1fil
%%%%%%%%%%%%%%%%%%%%%%%%%%%%%%%%%%%%%%%%%%%
\begin{IEEEbiography}[{\includegraphics[width=1in,height=1.25in,clip,keepaspectratio]{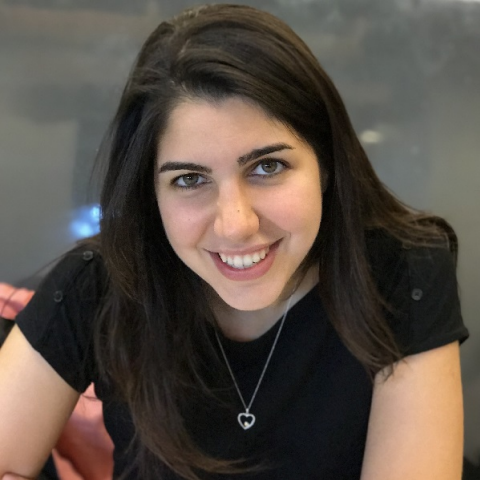}}]{Dorsa Sadigh}
is an Assistant Professor in the CS and EE departments at Stanford University. Her research interests lie in the intersection of robotics, learning and control theory. Specifically, she is interested in developing algorithms for safe and adaptive human-robot interaction. Dorsa has received her doctoral degree in Electrical Engineering and Computer Sciences (EECS) at UC Berkeley in 2017, and has received her B.Sc. in EECS at UC Berkeley in 2012.
\end{IEEEbiography}
%%%%%%%%%%%%%%%%%%%%%%%%%%%%%%%%%%%%%%%%%%%
\vskip -2.5\baselineskip plus -1fil
%%%%%%%%%%%%%%%%%%%%%%%%%%%%%%%%%%%%%%%%%%%
\begin{IEEEbiography}[{\includegraphics[width=1in,height=1.25in,clip,keepaspectratio]{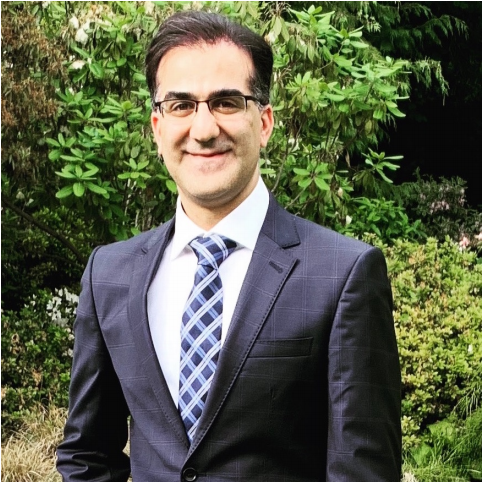}}]{Ramtin Pedarsani}
is an Assistant Professor in the ECE Department at the University of California, Santa Barbara. He received the B.Sc. degree in electrical engineering from the University of Tehran in 2009, the M.Sc. degree in communication systems from the Swiss Federal Institute of Technology (EPFL) in 2011, and his Ph.D. from the University of California, Berkeley, in 2015. His research interests include networks, game theory, machine learning, and transportation systems.
\end{IEEEbiography}
%%%%%%%%%%%%%%%%%%%%%%%%%%%%%%%%%%%%%%%%%%%
\vskip -2.5\baselineskip plus -1fil
%%%%%%%%%%%%%%%%%%%%%%%%%%%%%%%%%%%%%%%%%%%
\begin{IEEEbiography}[{\includegraphics[width=1in,height=1.25in,clip,keepaspectratio]{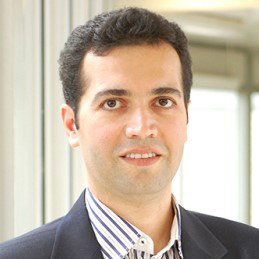}}]{Yaser P. Fallah} is an Associate Professor in the ECE Department at the University of Central Florida. He received the Ph.D. degree from the University of British Columbia, Vancouver, BC, Canada, in 2007. From 2008 to 2011, he was a Research Scientist with the Institute of Transportation Studies,
University of California Berkeley, Berkeley, CA, USA. His research, sponsored by industry, USDoT, and NSF, is focused on intelligent transportation systems and automated and networked vehicle safety systems.
\end{IEEEbiography}
%%%%%%%%%%%%%%%%%%%%%%%%%%%%%%%%%%%%%%%%%%%
% \vskip -2.5\baselineskip plus -1fil

\end{document}